%% file: acl_latex.tex
\pdfoutput=1

\documentclass[11pt]{article}

\usepackage[]{acl}

\usepackage{times}
\usepackage{latexsym}

\usepackage[T1]{fontenc}

\usepackage[utf8]{inputenc}

\usepackage{pifont}    
\usepackage{xcolor}    
\newcommand{\greentick}{{\color{green}\ding{51}}}   
\newcommand{\redcross}{{\color{red}\ding{55}}}      

\usepackage{microtype}
\usepackage{graphicx}
\usepackage{booktabs}
\usepackage{paralist}
\usepackage{listings}
\usepackage{todonotes}
\usepackage{float}

\usepackage{listings}
\usepackage{xcolor}

\usepackage{amsmath}
\usepackage{breqn}
\usepackage[most]{tcolorbox}
\usepackage{lipsum}
\usepackage{fvextra}

\title{UnSeenTimeQA: Time-Sensitive Question-Answering\\ Beyond LLMs' Memorization}

\author{
\textbf{Md Nayem Uddin}$^{\spadesuit}$\hspace{9pt}
\textbf{Amir Saeidi}$^{\spadesuit}$ \hspace{9pt}
\textbf{Divij Handa}$^{\spadesuit}$ \hspace{9pt}
\textbf{Agastya Seth}$^{\spadesuit}$ \hspace{9pt} \\
\textbf{Tran Cao Son}$^{\diamondsuit}$ \hspace{9pt}
\textbf{Eduardo Blanco}$^{\heartsuit}$ \hspace{9pt}
\textbf{Steven R. Corman}$^{\spadesuit}$ \hspace{9pt} 
\textbf{Chitta Baral}$^{\spadesuit}$ \vspace{2pt} \\
$^\spadesuit$Arizona State University \; $^\diamondsuit$ New Mexico State University \; $^\heartsuit$ University of Arizona  \vspace{2pt}\\
\texttt{muddin11@asu.edu}
}

\begin{document}
\maketitle
\begin{abstract}


This paper introduces \emph{UnSeenTimeQA}, 
a novel data contamination-free
time-sensitive question-answering (TSQA) benchmark.
It differs from existing TSQA benchmarks by 
avoiding web-searchable queries grounded in the 
real world. 
We present a series of time-sensitive event scenarios 
based on synthetically generated facts.
It requires large language models~(LLMs) to engage 
in genuine temporal reasoning 
without depending on the factual knowledge 
acquired during the pre-training phase.
Our data generation framework enables on‐demand 
generation of new samples, 
mitigating the risk of data leakage.
We designed three types of 
time-sensitive questions 
to test LLMs' temporal reasoning abilities 
over sequential and parallel event occurrences.
Our evaluation of five LLMs on synthetic fact-based TSQA 
reveals mixed results: while they perform well on simpler subsets, 
their overall performance remains inferior 
as compared to real world fact-based TSQA.
Error analysis 
indicates that LLMs face difficulties 
in reasoning over long-range event dependencies 
and parallel events.
\end{abstract}

\input{tex/1_introduction}
\input{tex/2_previous_work}
\input{tex/3_existing_benchmark}

\input{tex/4_new_benchmark}

\input{tex/5_experiments}
\input{tex/6_result_analysis}
\input{tex/7_error_analysis}
\input{tex/8_conclusion}

\section*{Limitations}
We have introduced a novel TSQA benchmark focusing on 
durations within a 24-hour range. 
While this focus offers a controlled environment for evaluation, 
it does not encompass longer temporal durations 
that extend beyond a single day. 
We anticipate that it will inspire 
other researchers to expand on longer time-sensitive scenarios
as it requires adding more planning domains distributed across seconds, minutes, hours, days, weeks, months, and years. 
Thus, we acknowledge this limitation with a positive outlook, 
considering it an opportunity for future advancements in the field. 
Also, the UnSeenTimeQA benchmark employs a template-based approach for event description generation which often risks predictability and limited variability. We addressed these concerns by using a mixture of templates. This strategy ensures that the generated text remains diverse and avoids a repetitive or monotonous appearance. 
Our benchmark currently includes time-sensitive questions for which we always expect answers. 
To enhance the challenge of our work, 
incorporating unanswerable time-sensitive questions could significantly improve the benchmark. 

 
We only work with the English language.
Working exclusively with the English language can limit 
the scope and applicability of research findings
for other languages.
We note that UnSeenTimeQA could be easily 
extended to other languages by replacing the event templates.


\section*{Ethical Considerations}
The authors state that this work is in accordance
with the ACL Code of Ethics and does not raise
ethical issues. AI assistants, specifically Grammarly and ChatGPT, were utilized to correct grammatical errors and restructure sentences.

\section*{Acknowledgments}
This research was supported by a grant from the U.S. Office of Naval Research (N00014-22-1-2596). We utilized the Azure OpenAI service for querying ChatGPT with
credits from Microsoft’s Accelerating Foundation
Models Research program. We also thank Research Computing (RC) 
at Arizona State University (ASU) for providing compute resources for the experiments.

\bibliography{anthology,custom}
\bibliographystyle{acl_natbib}

\appendix
\onecolumn
\section*{Appendix}
\input{tex/a_experiment_samples}
\input{tex/a_additional_results}
\input{tex/a_templates}
\input{tex/a_question_samples}

\input{tex/a_prompts}
\input{tex/a_model_details}
\input{tex/a_dataset_datacard}

\end{document}

%% file: tex/1_introduction.tex
\section{Introduction}
Time-sensitive question-answering (TSQA) involves responding to queries that require an understanding of time and events \cite{Chen2021ADF}. 
For example: 
\emph{Which football club did Leo Messi play for in 2010?} and 
\emph{Which football club did Leo Messi play for in 2023?}
are time-sensitive questions because they require 
extracting information 
based on the temporal anchor in the questions (i.e., \emph{2010, 2023}).
Also, answering more complex time-sensitive questions such as \emph{Which football club did Leo Messi play for after FC Barcelona?} demands reasoning
over multiple events' timestamps, 
durations, 
and how they are temporally related. 

\renewcommand{\thefootnote}{}
\footnotetext[1]{Our code and data are available at: \url{https://github.com/nurakib/UnSeenTimeQA}}
\footnotetext[1]{Huggingface endpoint: \url{https://huggingface.co/datasets/nurakib/UnSeenTimeQA}}
\renewcommand{\thefootnote}{\arabic{footnote}}

\begin{figure}[t!]
    \centering
    \includegraphics[width=\columnwidth]{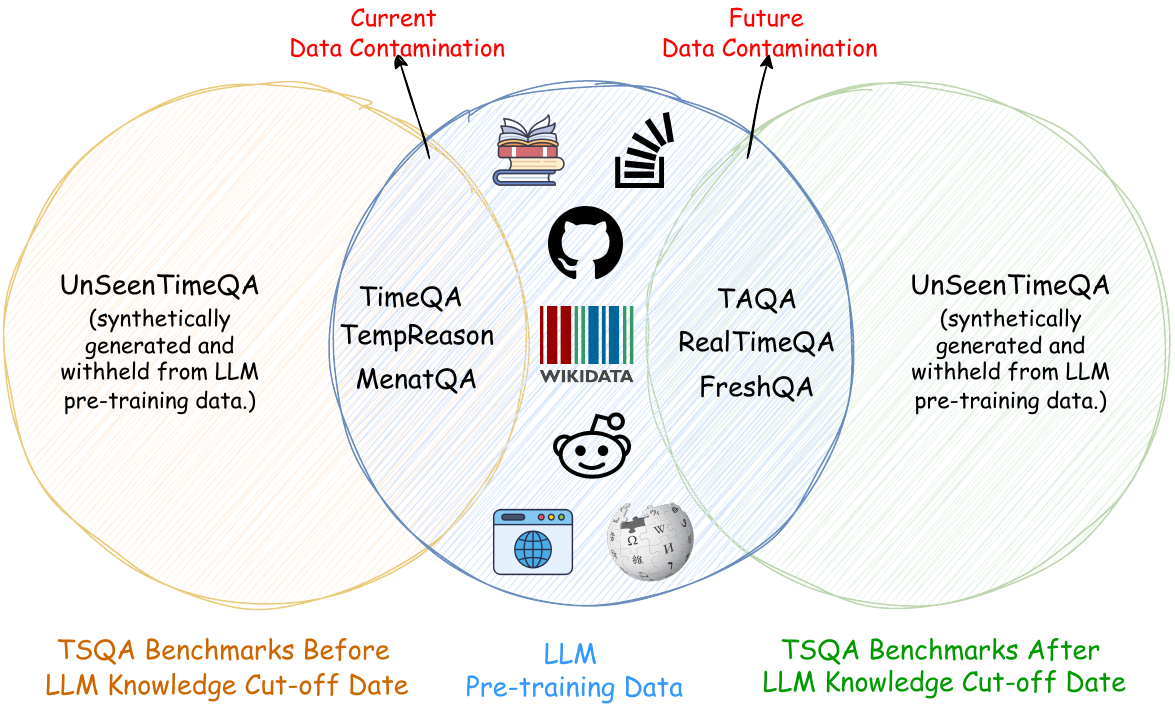}
    \caption{
    Schematic overview of UnSeenTimeQA with other time-sensitive question-answering benchmarks relative to LLMs' knowledge cut-off date. 
    Benchmarks developed 
    before the knowledge cut-off date 
    already suffer from data contamination 
    (yellow–blue overlap), 
    since their evaluation data was available on 
    the internet during 
    LLM pre-training phase. 
    Benchmarks developed 
    after the Knowledge cut-off date are at risk of 
    future contamination (green–blue overlap), 
    as newer pre-training corpora 
    may include their evaluation data. 
    In contrast, UnSeenTimeQA (non-overlapping with blue) utilizes synthetic facts, 
    providing a robust, contamination-free evaluation  
    to accurately measure temporal reasoning 
    capabilities of LLMs.
}
    \label{f:overview_tsqa}
\end{figure}

\begin{figure*}[!h]
    \centering
    \includegraphics[width=\textwidth]{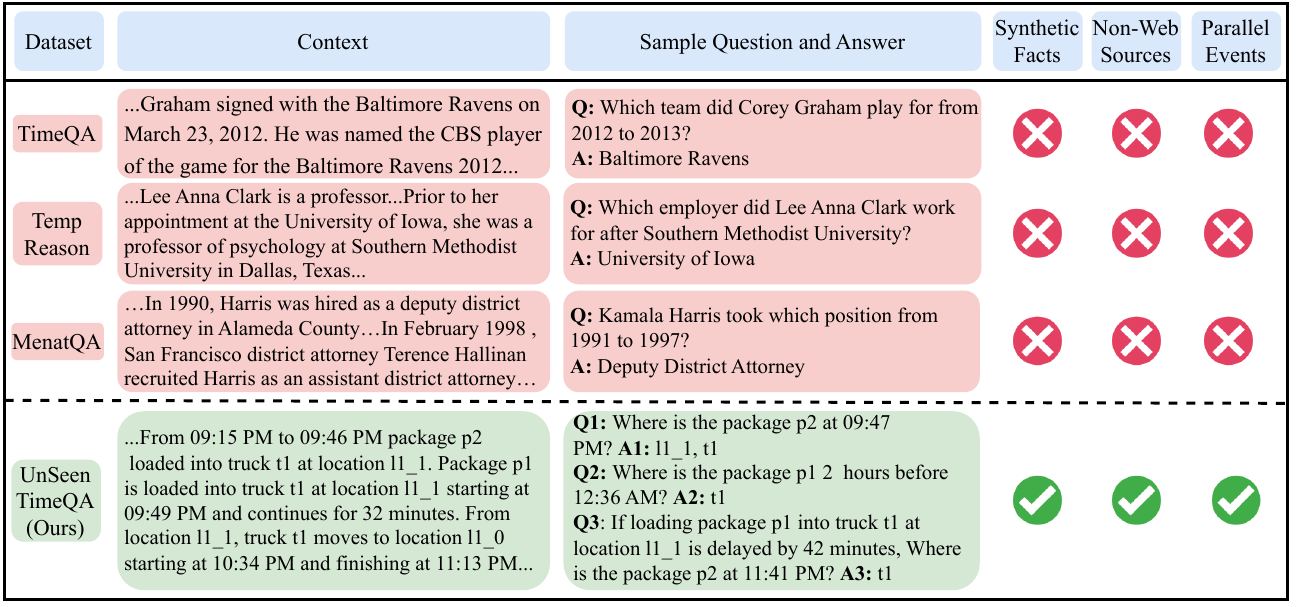}
    \caption{
    Examples from three TSQA benchmarks (top). These questions are based on real-world facts. Also, relevant information to answer these time-sensitive questions 
    exactly matches publicly available web sources. In contrast, we developed the UnSeenTimeQA benchmark (bottom), where the questions are based on synthetic facts and do not exactly align with any web sources. We also include parallel event scenarios for benchmarking TSQA. 
}
    \label{f:motivation}
\end{figure*}

TSQA benchmarks such as
TimeQA~\cite{Chen2021ADF},  
TempReason~\cite{tan-etal-2023-towards}, 
and MenatQA~\cite{wei-etal-2023-menatqa} 
primarily focus on answering
time-evolving factual questions
grounded in the real world. 
The questions are derived from Wikidata~\cite{vrandevcic2014wikidata}, 
and cover the employment histories
of individuals (e.g., athletes, politicians).
For instance, as Figure~\ref{f:motivation} highlights,
\emph{Which team did Corey
Graham play for from 2012
to 2013?} is a factual question that
can be answered by searching the web.
Since LLMs are pre-trained on vast text data from the web, 
including the entire Wikipedia,
using these benchmarks might not fully
test models' abilities for TSQA.
It is possible that LLMs 
rely on memorized facts acquired during 
the pre-training phase 
to answer time-sensitive questions~\cite{ dhingra-etal-2022-time, geva-etal-2023-dissecting, mallen-etal-2023-trust, fatemi2024test}
rather than demonstrating temporal reasoning.
RealTimeQA~\cite{kasai2023realtime}, 
FreshQA~\cite{vu-etal-2024-freshllms},
and TAQA~\cite{zhao-etal-2024-set}
are designed to avoid 
data contamination
by collecting 
questions after the LLMs knowledge cut-off date. 
These benchmarks require periodic manual updates.
If the manual updates are discontinued,
the data contamination issue will appear with
newly released LLMs.
For instance, RealtimeQA stopped its manual updates 
in January 2024---it is likely that 
new LLMs~(knowledge cut-off date after January 2024) will be exposed to the publicly available web sources related to RealtimeQA during 
the pre-training phase.
\newcite{meem-etal-2024-pat} proposed  
a self-updating TSQA benchmark
by creating present-anchored 
questions.
However, updates occur only after 
new information has been added to the Wikidata knowledge base.

In this work, 
we introduce UnSeenTimeQA—an evaluation benchmark for TSQA 
that avoids factual questions grounded in the real world and 
extends TSQA for complex parallel event occurrences.
Our approach eliminates the need for external knowledge bases or 
periodic manual annotations by treating 
events as synthetic facts. 
Because the event scenarios are generated synthetically, 
anyone can generate arbitrarily 
large number of new question on demand, 
ensuring that evaluation data never 
becomes stale or over-fitted.
This yields a truly data contamination–free 
evaluation for TSQA. 
As shown in Figure \ref{f:overview_tsqa}
UnSeenTimeQA entirely lies outside the overlap of present and future data contamination zones
when compared to six popular TSQA benchmarks, 
demonstrating its robustness against data contamination.

Additionally, 
existing TSQA benchmarks~\cite{suzgun-etal-2023-challenging}
fall short by treating events as sequential occurrences.
They ignore the scenarios 
where multiple events happen 
concurrently (e.g., the initiation of one event affects 
the outcome of other events).
While some benchmarks \cite{Chen2021ADF, su-etal-2024-living} include overlapping events (e.g., a person holding two jobs simultaneously), 
these are often simplistic as they are grounded 
in real-world facts.
Consider a complex reasoning scenario 
where two packages are simultaneously loaded into a truck. 
The overall completion time should be determined by 
the maximum duration among these two events.
This kind of reasoning is missing in existing
TSQA benchmarks.


To create the new benchmark, 
we draw inspiration from the logistics problems~\cite{long2000aips} 
of the International Planning Competitions (IPC). 
The resulting data samples are self-contained and 
only require temporal reasoning to determine the answers. 
We discuss the benchmark development process in Section \ref{s:new_banchmark}. 
Our major contributions are:

\begin{compactitem}
    \item Demonstrating that LLMs depend on memorized facts to answer time-sensitive questions from existing TSQA benchmarks~(Section~\ref{s:existing_benchmarks_sec}).
    \item Introducing UnSeenTimeQA---a data 
    contamination free benchmark to evaluate 
    LLMs' time-sensitive question-answering capabilities on
    sequential and parallel event occurrences.
    \item Allowing creation of new, contamination-free time-sensitive questions on demand via our open-source data generation framework.
    \item Evaluating five LLMs using UnSeenTimeQA and analyzing reasoning chains to find the
    most common error types.
    
\end{compactitem}


%% file: tex/2_previous_work.tex
\section{Previous Work}
Temporal reasoning enables 
NLP systems to interpret 
event sequences and their 
relationships as described in textual data. 
At its core, temporal reasoning involves determining event relations~\cite{ning-etal-2017-structured, ning-etal-2020-torque} and event ordering~\cite{cassidy-etal-2014-annotation, zhang-etal-2020-reasoning, zhou-etal-2022-generating}.
\newcite{chu-etal-2024-timebench} categorized temporal reasoning tasks into three main categories: 
symbolic, 
commonsense, and 
event temporal reasoning.
Symbolic temporal reasoning~\cite{tan-etal-2023-towards, thukral-etal-2021-probing} 
focuses on the comprehension 
of the abstract temporal expressions. 
Both temporal commonsense reasoning~\cite{zhou-etal-2019-going, zhou-etal-2021-temporal} and 
event temporal reasoning~\cite{Chen2021ADF} focus on 
understanding how events unfold over time
but the underlying tasks are different. 
Temporal commonsense reasoning is based 
on general knowledge and assumptions about the events, 
whereas event temporal reasoning focuses on specific facts, 
precise timing, and the order of events within a given context.
Time-sensitive question answering (TSQA) 
is an event temporal reasoning task. 
In TSQA, a system analyzes text 
to understand the temporal aspects of events. This includes identifying when events occur, how long they last, and their order. The goal is to answer questions accurately about event timing and relationships.

Initially, TSQA benchmarks~\cite{jia2018tequila, saxena-etal-2021-question} were developed 
using temporal knowledge bases. 
As a result, their scope remains limited to the entities 
within the knowledge bases. 
TimeQA~\cite{Chen2021ADF} is one of the largest benchmarks
to tackle time-sensitive questions derived from natural language. 
It requires significant annotation effort 
to align the Wikipedia knowledge base 
with corresponding articles. 
The questions in TimeQA require reasoning over event-time relationships. 
To enhance the complexity, \newcite{tan-etal-2023-towards} introduced TempReason, a benchmark that addresses both event-time and event-event relations. 
MenatQA~\cite{wei-etal-2023-menatqa} 
adds an additional layer of complexity 
by integrating counterfactual questions 
into the TSQA benchmarks. 
Also, TempTabQA~\cite{gupta-etal-2023-temptabqa} 
is developed to benchmark TSQA for semi-structured data. 
Regardless of the temporal coverage of these benchmarks,
solely relying on real-world facts for time-sensitive questions makes these benchmarks highly susceptible to data contamination due to the pre-training frameworks of LLMs. 
To tackle this, developing TSQA benchmarks~\cite{kasai2023realtime, vu-etal-2024-freshllms, zhao-etal-2024-set} after 
LLMs pre-training knowledge cut-off date is proposed. 
However, these benchmarks require frequent updates, 
making the approach unsustainable in the long run 
given the rapid release of new LLMs.

In contrast, we propose a new benchmark, 
drawing inspiration from complex planning problems. 
Our benchmark is resistant to data contamination because 
it does not rely on real-world facts 
to generate time-sensitive questions. 
Also, introducing parallel event execution to UnSeenTimeQA 
is novel as compared to any existing TSQA benchmarks.




%% file: tex/3_existing_benchmark.tex
\section{Do Existing TSQA Benchmarks Address Temporal Reasoning?}
\label{s:existing_benchmarks_sec}

Existing TSQA benchmarks include questions 
about real-world facts 
that might have been present in LLMs pre-training data.
For instance, 
questions such as \emph{``Who was the US President in 2008?''} 
can be answered by recalling LLMs' 
memorized knowledge rather than 
temporal reasoning.
To assess LLMs’ temporal reasoning ability, 
we evaluated six TSQA benchmarks, 
split into two groups.
The first group includes 
TSQA benchmarks
(i.e.,~\emph{TimeQA, TempReason, MenatQA}) 
developed before LLM knowledge cut-off date, 
which are highly likely to be present 
in LLMs pre-training data. 
The second group includes TSQA benchmarks developed after LLM knowledge cut-off date. 
These benchmarks (i.e.,~\emph{FreshQA, RealtimeQA, and TAQA})
are designed to tackle data contamination and less likely to be present in LLMs pre-training data. 
We randomly sampled 1500 questions from six benchmarks; 
see Appendix \ref{a:existing_benchmark} for more details.

\begin{table}
\small
\centering
\resizebox{\columnwidth}{!}{
\input{tables/results_on_prior_benchmark_2}
}
\caption{Results with three TSQA datasets---TimeQA, TempReason, and MenatQA---under four conditions: 
no context~(\textbf{w/o C}), 
gold context~(\textbf{w/ GC}), 
altered context~(\textbf{w/ AC}), 
and altered context and questions~(\textbf{w/ ACQ}). 
The answer accuracy of the GPT-4 model is similar answering time-sensitive questions (a) without contexts (w/o C) and (b) altering the gold contexts and the questions (w/ ACQ). 
This indicates the GPT-4 model answers the time-sensitive questions based on memorized facts rather than using temporal reasoning and information from the provided context. 
For the MenatQA (counterfactual) split, context alteration is not possible because their contexts do not include the gold answers.
}

\label{t:results_existing_benchmarks}
\end{table}

\subsection{TSQA Benchmarks Before LLMs Knowledge Cut-off Date}
\label{ss:existing_benchmarks_sec_grp_1}
In TSQA benchmarks developed before the LLM knowledge cut-off date, 
each question is paired with a gold context. 
The gold context (i.e., the relevant Wikipedia article) 
is assumed to contain all the necessary information 
to answer the question correctly.
We hypothesize that 
if an LLM primarily depends on its memorized factual knowledge 
to answer time-sensitive questions, 
then its performance will drop 
when presented with altered versions of the gold context.
This is because such alterations would introduce information 
that is not encountered during the LLM's pre-training phase, 
creating a conflict between memorized facts. 
For the empirical analysis, we evaluated 
a GPT-4 model under the following conditions:

\noindent \textbf{Without Context}: 
We initially set a baseline by asking 
the time-sensitive questions without
providing any context, meaning the model
can only answer the time-sensitive questions 
based on the knowledge acquired 
at the pre-training phase. \\
\noindent \textbf{With Gold Context}:
Subsequently, we evaluated the model's performance 
when provided with the gold context alongside the questions.\\
\noindent \textbf{With Altered Context:} 
We evaluated the model's performance on time-sensitive questions with an altered context 
where the gold answers were replaced with 
plausible but incorrect random entities. 
The question itself remained unchanged. 
For instance, consider the question: ``Where did Leo Messi play in 2010?''. 
If the original gold context stated that ``Lionel Messi played for FC Barcelona in 2010'' (making FC Barcelona the initial correct answer), 
the altered context might state ``Lionel Messi played for FC Aftermath in 2010'' 
(making FC Aftermath the correct answer according to the altered context). 
This alternation setup evaluated the model's ability to prioritize 
the altered context over memorized facts.\\
\noindent \textbf{With Altered Context and Question:} 
In this condition, we start with the altered context from the previous stage.
We additionally alter the main entity in the questions and the gold contexts.
For example, if the original question was ``Where did Leo Messi play in 2010?'', 
it was changed to ``Where did Teo Tsiuri play in 2010?''. 
Consequently, the main entity in the gold context, ``Lionel Messi'', 
was replaced with ``Teo Tsiuri''. 
This setup was designed to force the model 
to reason with entirely novel entities,
as its internal knowledge base would not contain information about these new entities. 
We replace entities by exact matches in the gold context, 
which may lead to inconsistencies in some samples. 
However, our manual analysis shows these instances are negligible.

Results in Table \ref{t:results_existing_benchmarks}
show the GPT-4 model can correctly answer 33\% to 44\% 
of time-sensitive questions from the existing benchmarks 
without providing the gold context. 
Adding the gold context in the prompt allows 
the model to answer more accurately. 
However, altering the answers with random entities
in the gold context leads to lower performance.
Altering entities in both contexts and questions reduces 
the model's performance almost to the level of 
asking questions without providing any context.
This performance drop across all three benchmarks hints 
at possible memorization of facts
while answering time-sensitive questions.



We also engage three human reviewers to find web articles that precisely match the answers to the time-sensitive questions. 
Reviewers are limited to browsing five articles per question.
If the reviewers are able to find time-sensitive facts 
in publicly available web articles, 
it indicates a high likelihood that 
the LLMs were exposed to these articles
during pre-training. 
We found that in
88\%-98.8\% of cases reviewers found 
answers to the time-sensitive questions 
using Google searches. 
The task for reviewers involves no reasoning; 
they simply find answers directly from the web. 
Similarly, when LLMs generate answers, 
they might recall 
memorized knowledge acquired during the pre-training phase, 
rather than engaging in temporal reasoning.

\begin{figure}[t!]
    \centering
    \includegraphics[width=\columnwidth]{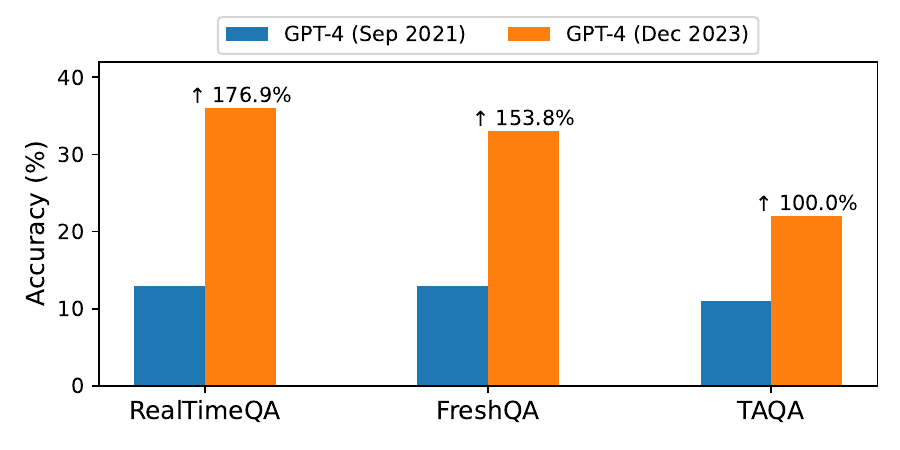}
    \caption{
    Results with three recent TSQA benchmarks.
    We evaluate the same LLM 
    with different knowledge cut-offs. 
    Since the questions are anchored in 2023, 
    the old GPT-4 model, with knowledge upto \emph{September 2021}, 
    shows lower performance. 
    In contrast, the new GPT-4 model, 
    with knowledge updated until \emph{December 2023}, 
    answers the questions with much higher accuracy, 
    ranging from 100\% to 177\% increase.
}
    \label{f:existing_benchmark_2}
\end{figure}

\subsection{TSQA Benchmarks After LLMs Knowledge Cut-off Date} 
\label{ss:existing_benchmarks_sec_grp_2}
Recent efforts to mitigate 
data contamination 
involve developing 
TSQA benchmarks~(\emph{i.e., FreshQA, RealtimeQA, and TAQA})
after the LLMs pre-training knowledge cut-off date.
However, these benchmarks rely  
on continuous manual updates to remain effective. 
If this manual updating stops 
(for instance, RealtimeQA stopped updating questions 
in January 2024),
these benchmarks can be easily contaminated
because newer LLMs with updated knowledge cut-offs 
may have been exposed 
to these time-sensitive real-world facts during their pre-training phase.

To empirically evaluate this, 
we randomly sampled 150 questions and  
their answers (as of 2023) from each benchmark. 
We asked these questions (without context)
to two different versions of a GPT-4\footnote{
Since the architectural details of the GPT-4 models are not publicly available, 
we assume that the knowledge cut-off date is the primary differentiator
between the two models.} 
model. 
Results in Figure \ref{f:existing_benchmark_2} 
indicate that the newer GPT-4 model 
with updated knowledge cut-off date up to \emph{December 2023}
can answer time-sensitive questions using pre-trained knowledge 
as compared to the older GPT-4 model with knowledge cut-off 
date back to \emph{September 2021}. 
Specifically, we observed improvements ranging from 
100\% to 177\% when compared to the old GPT-4 version. 
So, manually updating  
the contamination-free TSQA benchmarks
is a never-ending process
and not sustainable in the long term.

While existing benchmarks have provided a starting point, 
their reliance on real-world facts undermines TSQA evaluation. 
Based on our analyses, a more reliable TSQA benchmark
is needed to avoid data contamination 
and the need for frequent manual updates 
for time-sensitive questions.

%% file: tables/results_on_prior_benchmark_2.tex
\begin{tabular}{lcccc}
\toprule
\textbf{Dataset} & \textbf{w/o C} & \textbf{w/ GC} & \textbf{w/ AC} & \textbf{w/ ACQ} \\
\midrule
\textbf{TimeQA} & & & & \\
~~Easy (150)               & 44\%          & 74\%                  & 52\%                                    & 46\%      \\ 
~~Hard (150)               & 39\%          & 71\%                  & 56\%                                   & 37\%      \\
\midrule
\textbf{TempReason}    & & & & \\
~~Event-Time (150)                & 40\%           & 66\%                  & 28\%                                   & 32\%      \\ 
~~Event-Event (150)                & 35\%           & 69\%                  & 40\%                                    & 37\%      \\
\midrule
\textbf{MenatQA}       & & & & \\
~~Scope (150)              & 39\%           & 80\%                  & 53\%                                    & 41\%      \\ 
~~Order (150)             & 35\%           & 75\%                  & 57\%                                    & 43\%      \\ 
~~Counterfactual (150)     & 33\%           & 54\%                  & N/A                                    & N/A       \\ 
\bottomrule
\end{tabular}

%% file: tex/4_new_benchmark.tex
\section{UnSeenTimeQA Benchmark}
\label{s:new_banchmark}

\subsection{Data Source}
We choose the logistics domain~\cite{long2000aips} 
from the International Planning Competition (IPC) 
to develop the UnSeenTimeQA benchmark. 
This domain includes multiple cities, 
and each city has certain locations. 
The planning task requires 
transporting packages 
from their initial locations 
to specified destinations. 
Within this predefined environment, 
six distinct events can occur. These events are: 
\emph{load truck},
\emph{unload truck},
\emph{drive truck},
\emph{load airplane},
\emph{unload airplane},
and \emph{fly airplane}.

A plan solver must create 
a valid sequence of events 
that relocates each package 
from its initial location 
to its designated destination.
We take these valid event sequences
and add a random duration to each event 
to make time-sensitive scenarios. 
For instance, consider a scenario 
where loading a product into a truck 
takes place at a specific location for 30 minutes. Afterward, the truck travels to another location, 
taking 50 minutes. 
Adding duration to each event
helps to incorporate 
the time-sensitive nature of 
the product's location, which changes over time. 

The valid plans are generated 
using the data generation pipeline from ActionReasoningBench~\cite{handa2024actionreasoningbench}. 
The events in the valid plans 
are translated into natural language descriptions. 
We use multiple templates to ensure linguistic variety 
and avoid uniformity. 
See the templates in Appendix~\ref{a:templates}.
In total, we employ 10 planning scenarios, 
each containing between 25 and 33 events.

\begin{figure}[!t]
    \centering
    \includegraphics[width=\columnwidth]{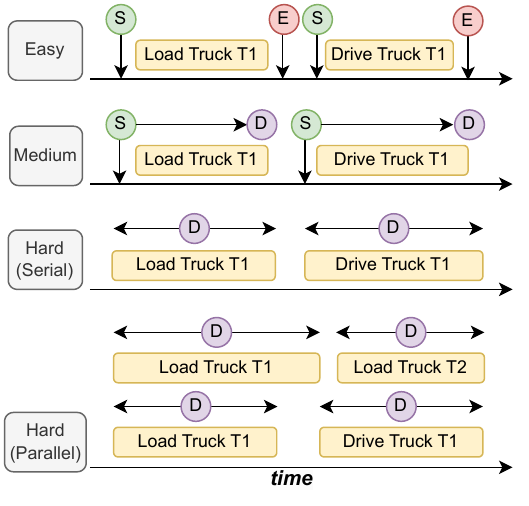}
    \caption{
    Example events from the UnSeenTimeQA benchmark. The benchmark is structured into four difficulty levels: easy, medium, hard (serial), and hard (parallel). In the easy level, the start (S) and end (E) times of each event are given. The medium level includes the start time (S) and duration (D) of each event. The hard (serial) level presents only the duration (D) of events, assuming sequential event occurrence. The hard (parallel) level also includes only durations (D), but events can occur simultaneously (Two products are loaded in the Truck T1 before starting the driving). 
}
    \label{f:new_benchmark}
\end{figure}

\subsection{Event Temporal Information}
Temporal information specifies when events occur by providing timestamps and durations. 
Based on the temporal information provided for each event, 
we categorize the data samples into three difficulty levels: \emph{Easy, Medium, and Hard}.
Figure \ref{f:new_benchmark} shows illustrative examples
for all three difficulty levels. 

\noindent \textbf{Easy:} 
Events are marked by a start and end timestamps~(\emph{e.g., 08:11 AM to 08:54 PM}). Such temporal information facilitates straightforward interpretations of the event order.

\noindent \textbf{Medium:} Events are marked by a start time and a duration (\emph{e.g., 08:11 AM, 43 minutes}). This format indicates when the events begin, but requires models to infer the end times.

\noindent \textbf{Hard:} Events are marked by a duration 
only (\emph{e.g., 43 minutes}). 
The absence of any specific indicator of 
the event start or end times 
adds complexity to determining the order of events. 
Moreover, this format facilitates the parallel event execution 
in addition to the existing sequential event execution.

\subsection{Question Types}
All the questions in the UnSeenTimeQA benchmark
focus on determining the locations of the packages
at different times. These questions are categorized 
into three groups, based on the temporal modifiers.

\noindent \textbf{Static-Time:} This type of question involve asking about the location of a package at a specific time. For example, given a list of events occurring at different times, a static-time query is: \emph{``Where is package p0 at 10:53 PM?''}

\noindent \textbf{Relative-Time:} This type of questions involve asking about the location of a package modified by a certain temporal offset from a specified time. For example, given a list of events occurring at different times, a relative-time query is: \emph{``Where is package p0 2 hours after 8:13 PM?''}

\noindent \textbf{Hypothetical-Time:} This type of question involves creating hypothetical scenarios by altering the duration of an event in the given sequence. This approach forces the model to reason over a trajectory that differs from the narrated sequence of events. For example, after presenting a list of events occurring at specific times, a hypothetical-time question is: \emph{``If driving truck t1 from location l1\_1 to location l1\_0 is delayed by 66 minutes, Where is the product p0 at 10:18 PM?''} Answering such questions requires the model to anticipate on possible outcomes and the implications of timing changes on the overall event trajectory.


\begin{table*}[t]
\small
\centering
\resizebox{\textwidth}{!}{
\input{tables/results_on_easy_and_mid}
}

\caption{
Average accuracy and standard deviation across three splits of the easy and medium-level questions from UnSeenTimeQA for five LLMs. Each split contains 300 questions per question type. Larger models (Llama-3.1-70B, GPT-4o) outperform smaller models, highlighting the difficulty smaller models have in handling simple time-sensitive questions. In contrast, human evaluation (average accuracy among three reviewers) on a smaller subset (15 questions per difficulty level) achieves perfect accuracy.
}

\label{t:unseen_benchmarks_1}
\end{table*}

\begin{table*}[t]
\small
\centering
\resizebox{\textwidth}{!}{
\input{tables/results_on_hard}

}


\caption{
Average accuracy and standard deviation across three splits of the hard serial and hard parallel level questions from UnSeenTimeQA for five LLMs. Each split contains 300 questions per question type. Both hard serial and hard parallel questions pose substantial challenges for all models evaluated, with even the best-performing GPT-4o model incorrectly answering more than half of the hard parallel questions. Human evaluation (average accuracy among three reviewers on a subset of 15 questions per difficulty level) demonstrates significantly higher accuracy than the LLMs, showing a notable performance gap on these challenging questions.
}
\label{t:unseen_benchmarks_2}
\end{table*}

\subsection{Questions and Answers Generation}
We have developed an event tracking system
to determine the linked events 
for each package from 
the initial pickup to the final delivery.
Linked events are the events where a package is present—whether 
it is being loaded, unloaded, 
or transported.
For instance, if a product is picked up in event 1 
and reaches its destination in event 10,
our event tracking system will build a timeline of 
linked events for the package 
from event 1 to event 10.
From the timeline, we know exactly where the package is at any moment. 
We automatically generate 
time-sensitive questions from this evolving location information over time.
For answer generation, we consider two scenarios: 
if a package is in transit, 
the location of the package is designated 
to the vehicle transporting it. 
Alternatively, if the package is being loaded or unloaded, 
both the physical location and the vehicle involved in the event are correct answers. 
This explicit assumption 
is stated in the domain description
to make 
answer generation unambiguous.
We conducted several stages of quality checks to ensure
the accuracy of the generated timelines. 
This process was repeated until all identified issues were resolved.

Each sample in the UnSeenTimeQA benchmark 
follows a consistent data structure. 
Initially, we provide a domain description, 
which offers a description of the logistics problem. 
This description serves as a guide to the model 
about the possible events in the logistics environment. 
Following this, we detail the objects available in the environment and their initial states. 
Then we have a sequence of events occurring at different timestamps followed by a time-sensitive question. Example questions and answers are in Appendix \ref{a_example_samples}.

\subsection{Data Statistics}
We categorize temporal information 
into three difficulty levels: \emph{Easy, Medium, and Hard}. 
The hard category is further divided into 
\emph{Serial} and \emph{Parallel} event execution types. 
So, there are four variations: 
Easy (Serial), 
Medium (Serial), 
Hard (Serial), 
and Hard (Parallel).
To account for the complexity of the questions, 
we define a depth metric. 
It is the absolute positional difference between the start and end events relevant to a question 
in the given event sequence. 
A question with higher depth entails more complexity 
to the question. We consider depths ranging from 6 to 20 events 
for the \emph{UnSeenTimeQA} benchmark.
We randomly sample 20 questions for each depth, a total of 300 questions for each split. This process is repeated three times, resulting in three data splits for each question type across the four difficulty levels. 
In total, \emph{UnSeenTimeQA} encompass 10,800 data samples.

%% file: tables/results_on_easy_and_mid.tex
\begin{tabular}{l cccc|cccc} 
\toprule \textbf{Model} & \multicolumn{4}{c}{\textbf{Easy}} & \multicolumn{4}{c}{\textbf{Medium}} \\ 
\cmidrule{2-5} \cmidrule{6-9} & Static-Time & Relative-Time & Hypothetical-Time & Average & Static-Time & Relative-Time & Hypothetical-Time & Average\\ \midrule 

Gemma-2-9B & 79.11\scriptsize{\(\pm\)3.67} & 59.66\scriptsize{\(\pm\)1.22} & 45.55\scriptsize{\(\pm\)3.86} & 61.44 & 79.55\scriptsize{\(\pm\)1.57} & 60.22\scriptsize{\(\pm\)2.52} & 43.22\scriptsize{\(\pm\)3.00} & 61.00 \\ 
Gemma-2-27B & 75.22\scriptsize{\(\pm\)1.83} & 67.66\scriptsize{\(\pm\)1.52} & 57.88\scriptsize{\(\pm\)3.59} & 66.92 & 71.77\scriptsize{\(\pm\)1.26} & 68.00\scriptsize{\(\pm\)7.83} & 51.33\scriptsize{\(\pm\)2.30} & 63.70 \\ 
Llama-3.1-8B & 75.77\scriptsize{\(\pm\)3.33} & 45.00\scriptsize{\(\pm\)1.00} & 49.00\scriptsize{\(\pm\)1.45} & 56.59 & 70.44\scriptsize{\(\pm\)0.50} & 36.44\scriptsize{\(\pm\)5.33} & 48.77\scriptsize{\(\pm\)5.27} & 51.88 \\ 
Llama-3.1-70B & 97.00\scriptsize{\(\pm\)0.66} & 95.33\scriptsize{\(\pm\)1.76} & 85.55\scriptsize{\(\pm\)1.34} & 92.62 & 97.44\scriptsize{\(\pm\)0.50} & 88.33\scriptsize{\(\pm\)1.76} & 83.88\scriptsize{\(\pm\)2.83} & 89.88 \\  
GPT-4o & 96.33\scriptsize{\(\pm\)1.52} & 94.55\scriptsize{\(\pm\)2.14} & 90.11\scriptsize{\(\pm\)1.50} & 93.66 & 96.66\scriptsize{\(\pm\)2.33} & 92.77\scriptsize{\(\pm\)2.14} & 89.33\scriptsize{\(\pm\)2.40} & 92.92 \\ \midrule
Human & 100 & 100 & 100 & 100 & 100 & 100 & 100 & 100 \\ \bottomrule

\end{tabular}

%% file: tables/results_on_hard.tex
\begin{tabular}{l cccc|cccc} 
\toprule \textbf{Model} & \multicolumn{4}{c}{\textbf{Hard (Serial)}} & \multicolumn{4}{c}{\textbf{Hard (Parallel)}} \\ 
\cmidrule{2-5} \cmidrule{6-9} & Static-Time & Relative-Time & Hypothetical-Time & Average & Static-Time & Relative-Time & Hypothetical-Time & Average\\ \midrule 

Gemma-2-9B & 18.44\scriptsize{\(\pm\)1.83} & 15.55\scriptsize{\(\pm\)2.67} & 20.77\scriptsize{\(\pm\)1.83} & 18.25 & 16.22\scriptsize{\(\pm\)0.69} & 11.44\scriptsize{\(\pm\)2.03} & 17.33\scriptsize{\(\pm\)1.67} & 15.00 \\
Gemma-2-27B & 13.00\scriptsize{\(\pm\)1.85} & 14.66\scriptsize{\(\pm\)0.66} & 17.77\scriptsize{\(\pm\)0.77} & 15.14 & 12.99\scriptsize{\(\pm\)1.20} & 12.99\scriptsize{\(\pm\)2.90} & 15.10\scriptsize{\(\pm\)2.34} & 13.69 \\  
Llama-3.1-8B & 24.33\scriptsize{\(\pm\)1.85} & 23.00\scriptsize{\(\pm\)1.52} & 21.33\scriptsize{\(\pm\)1.73} & 22.88 & 22.77\scriptsize{\(\pm\)0.38} & 17.66\scriptsize{\(\pm\)1.76} & 22.55\scriptsize{\(\pm\)2.03} & 21.98 \\ 
Llama-3.1-70B & 41.50\scriptsize{\(\pm\)1.64} & 40.00\scriptsize{\(\pm\)0.47} & 33.66\scriptsize{\(\pm\)0.94} & 38.38 & 42.50\scriptsize{\(\pm\)2.12} & 36.16\scriptsize{\(\pm\)2.12} & 40.33\scriptsize{\(\pm\)0.94} & 39.66 \\  

GPT-4o & 57.11\scriptsize{\(\pm\)1.57} & 47.44\scriptsize{\(\pm\)2.87} & 44.77\scriptsize{\(\pm\)3.01} & 49.77 & 47.33\scriptsize{\(\pm\)2.60} & 39.11\scriptsize{\(\pm\)2.98} & 42.11\scriptsize{\(\pm\)1.83} & 42.85 \\ 
\midrule 
Human & 100 & 93.33 & 86.66 & 93.33 & 93.33 & 86.66 & 73.33 & 84.44 \\ \bottomrule

\end{tabular}

%% file: tex/5_experiments.tex
\section{Experiments}
\label{s:experiments}
\paragraph{Models:} We use 
zero-shot chain-of-thought~\cite{kojima2022large} 
prompting to evaluate LLMs' performance without any external influences
when answering time-sensitive questions.
Each prompt features a domain description, 
object description, initial states description, an event sequence, and a question. 
We add a formatting instruction at the end of the prompt. The instruction defines the response structure, including reasoning steps and the final answer.
Refer to Appendix \ref{a:experiment_prompt} for the sample prompt. 
All the experiments are conducted on five different LLMs. 
We include both open-weight and closed-weight LLMs variants. 
Among the open-weight models, we chose Llama-3.1-8B-Instruct, Llama-3.1-70B-Instruct~\cite{dubey2024llama}, Gemma-2-8B-it, and Gemma-2-27B-it~\cite{team2024gemma}.
The only closed-weight model is
GPT-4o~\cite{achiam2023gpt}. 
Model specific endpoints are listed in Appendix \ref{a:model_details}. 
We also conducted the same experiments in few-shot setup and report the results in Appendix \ref{a:few_shot_results}. 

\paragraph{Evaluation:} 
We report accuracy scores as model evaluation metric. 
LLM responses are split into two parts: 
1) Reasoning steps and 2) Final answer. 
If the correct answer is found in the final answer, 
it is considered correct; otherwise, incorrect.


%% file: tex/6_result_analysis.tex
\section{Results}

\paragraph{Performance Drops as the Difficulty Increases:}
The difficulty levels 
in UnSeenTimeQA 
are defined based on 
the event temporal information. 
We hypothesize that
LLMs will show a decreasing trend in accuracy 
as the difficulty levels increase.
Tables \ref{t:unseen_benchmarks_1} and \ref{t:unseen_benchmarks_2} 
show the overall performance 
drop with increasing difficulty. 
While the accuracy drop difference between 
easy and medium difficulty is marginal
(in some specific cases, medium is better);
the performance drop for 
the hard difficulty is higher. 
On average, we observe a 62\% accuracy drop 
from easy to hard (serial) difficulty and
a 65\% accuracy drop from easy to hard (parallel) difficulty. 
The accuracy drop remains nearly similar for
medium to hard (serial) 
and hard (parallel) transitions.
Overall, LLMs performed higher 
in scenarios (easy and medium)
where the task involves answering time-sensitive questions
given all explicit event timestamps and durations.
This points to models' strength in 
extraction-based tasks where explicit 
information is readily available. 
In contrast, the drop in accuracy 
for the both hard difficulty levels 
(where the models are required to deduce event timings only from durations)
points towards a potential
weakness in reasoning intensive scenarios. 
It is likely that tasks demanding 
more abstract reasoning from less direct information 
pose greater challenges for these models.


\begin{figure}[t!]
    \centering
    \includegraphics[width=\columnwidth]{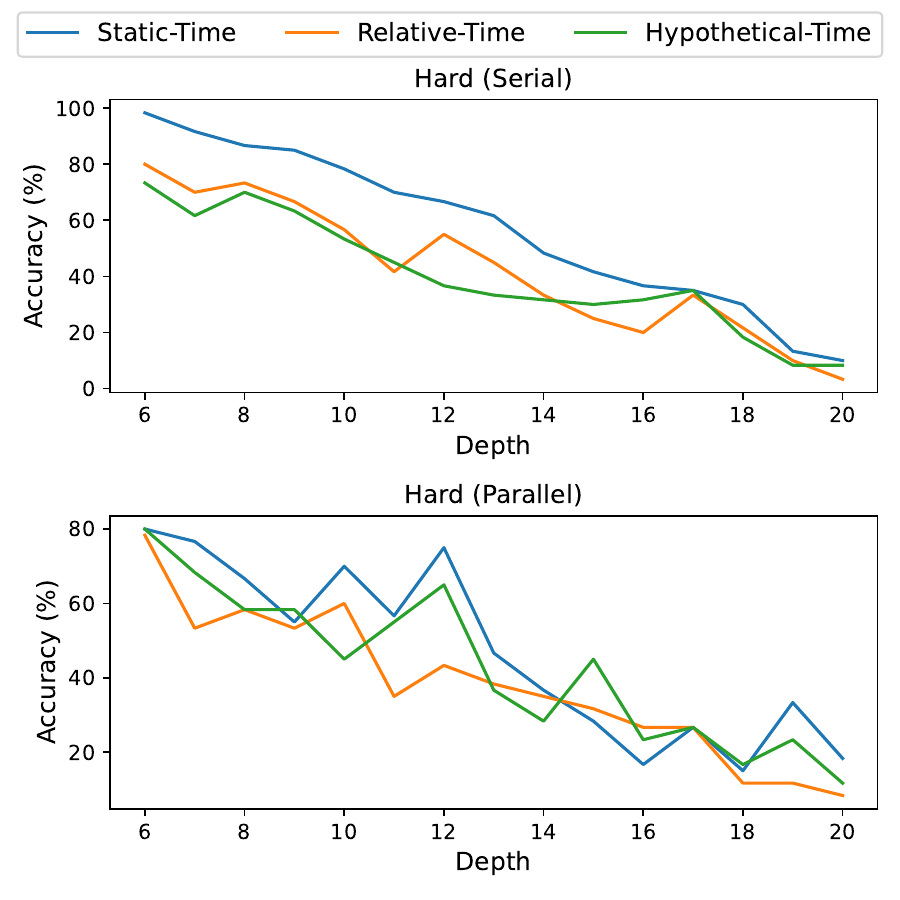}
    \caption{
    The depth-wise accuracy of the best-performing model (GPT-4o). Hard serial questions show
    a consistent decline in accuracy 
    with increasing depth. The accuracy trend 
    for the hard parallel questions is inconsistent. This suggests depth is not a reliable indicator
    for the hard parallel 
    questions complexity.
}
    \label{f:accuracy_depth}
\end{figure}

\paragraph{Performance across Question Types:}
We have three different types of time-sensitive questions.
Results in Table \ref{t:unseen_benchmarks_1} and Table  \ref{t:unseen_benchmarks_2} 
show most of the LLMs' performance 
decreases for the easy, medium, and hard (serial) difficulty levels 
when question complexity increases.
This suggests models make more errors in answering hypothetical-time questions 
(where the reasoning process involves iterating over an event narrative 
different from the given narrative).
However, we observe a different trend for question types in hard (parallel) difficulty.
The result indicates the relative-time questions are harder to answer as compared to static-time and hypothetical-time questions.

\begin{figure*}[h!]
    \centering
    \includegraphics[width=0.95\textwidth]{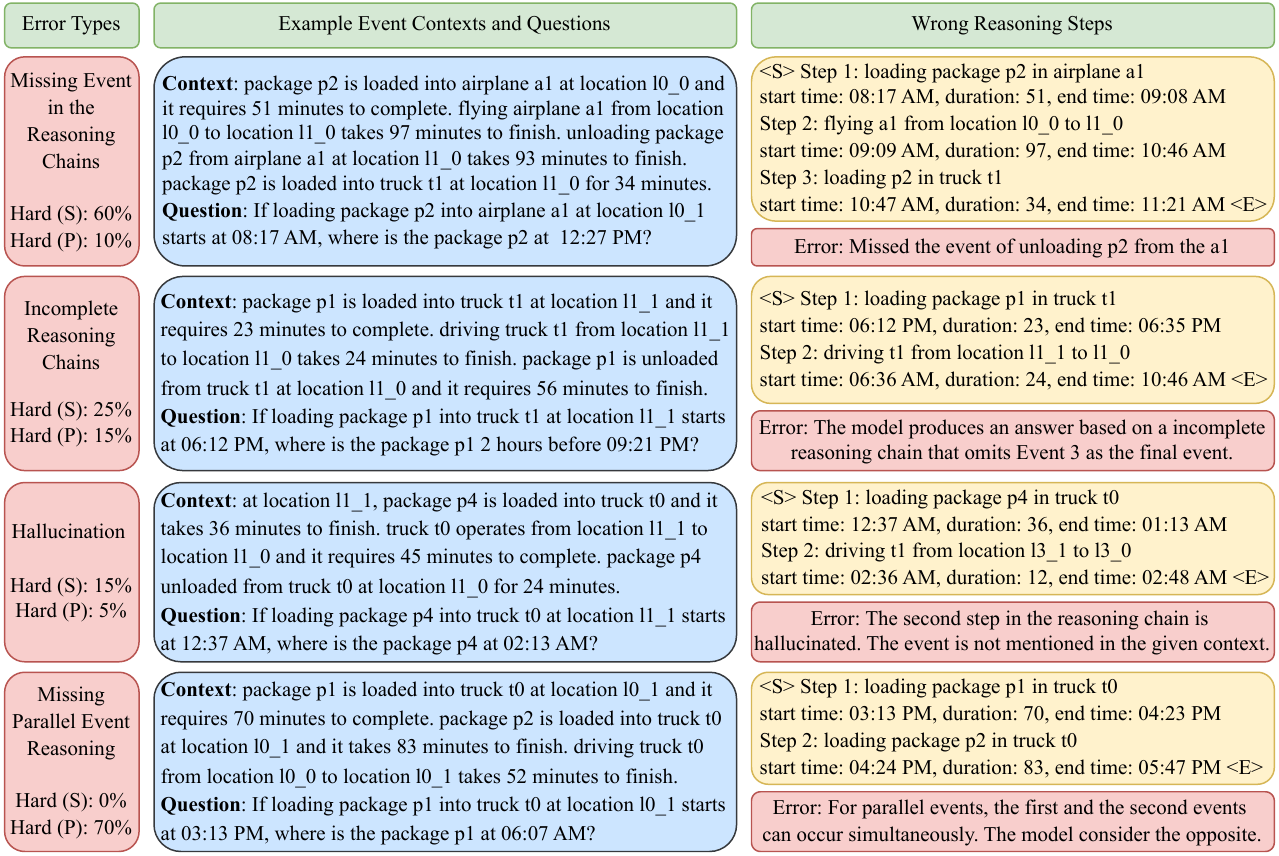}
    \caption{
     Error types identified in the best-performing model predictions (GPT-4o). For hard serial, GPT-4o often misses relevant events in  reasoning chain and for hard parallel, the model struggles with parallel events. 
}   
    \label{f:unseen_errors}
\end{figure*}

\paragraph{Performance across Question Depth:}
Question depth is a complexity indicator 
for time-sensitive questions. 
It refers to the number of events 
between the start to the target event time.
For example, 
\emph{``Where is product p1 at 10:20 PM?''}, 
the depth is determined by the total events occurring 
from the start time up to 10:20 PM.
Higher depth means reasoning 
over more events to find the correct answers.
Figure \ref{f:accuracy_depth} shows 
depth-wise accuracy for all three question types 
in hard difficulty for the best-performing model (\emph{GPT-4o}).
For hard serial question, we observe a consistent 
decrease in accuracy as 
depth increases for all question types.
This indicates depth is a reliable 
complexity metric for hard serial questions. 
The accuracy trend for hard parallel 
questions are also decreasing but inconsistent across depth, 
suggesting that depth alone 
may not be a reliable complexity indicator. 
This is plausible because, in parallel questions, 
depth can be higher, 
but the best reasoning chain is shorter 
due to the parallel executability of events.
Therefore, alternative metrics may be needed to better 
capture the complexity of parallel time-sensitive questions.
We also observe a similar trend of depth for the second-best model (\emph{Llama-3.1-70B}). 


%% file: tex/7_error_analysis.tex
\section{Error Analysis}
We conduct an error analysis of the GPT-4o (best-performing model) predictions, 
focusing on step-by-step reasoning chains. 
We randomly sampled 60 errors from hard serial and hard parallel questions. 
We ignored the easy and medium splits for error analysis
as those splits are less challenging for LLMs.
We manually identified four common error types, 
summarized in Figure \ref{f:unseen_errors}. 
In cases where multiple errors occur within a reasoning chain, 
we categorized the error type based on the first error encountered 
to maintain consistency.

The most common error type for hard serial questions
is \textbf{\emph{missing an event in the reasoning chain}}, 
accounting for 60\% of the errors. 
This occurs primarily when the model needs to track 
long-range event dependencies but fails to account 
for a relevant event in the reasoning chain. 
Another error type
is \textbf{\emph{incomplete reasoning chains}} (25\%).
This type of error particularly appears in relative-time questions. 
These questions use before and after comparators
in the questions, 
and the model prematurely shifts focus to reasoning 
about the final time, leaving the reasoning chain incomplete. 
We also observed \textbf{\emph{hallucinations}} in some cases (15\%), where the model generated events or objects not present in the context.
For hard parallel questions, 
hallucinations are less, only 5\%.
However, a unique error type 
emerged---\textbf{\emph{missing parallel event reasoning}}.
This error type points to the model shortcomings in reasoning 
over parallel events. 
For instance, when two packages are simultaneously loaded 
onto the same truck, the model often misses the 
parallel aspect and treats these two loading events 
as sequential occurrences. 
In 70\% of the hard parallel errors, 
the model failed to account for events that occurred in parallel, which differentiates this from the missing events error in hard serial questions. Incomplete reasoning chains are also observed in 15\% of the hard parallel errors.

%% file: tex/8_conclusion.tex
\section{Conclusion}

In this work, we introduced UnSeenTimeQA, a novel benchmark designed to evaluate the temporal reasoning capabilities of LLMs. Unlike existing time-sensitive question-answering (TSQA) benchmarks, UnSeenTimeQA eliminates the reliance on real-world factual knowledge. Our findings highlight a critical shortcoming in current LLMs: their dependence on pre-trained knowledge for answering time-sensitive questions. By focusing on synthetic and controlled data generation environments, we ensure that answers cannot be pre-learned or leaked to the LLM. This approach allows us to more accurately measure LLMs ability to reason about time-sensitive information that reflects real-world reasoning demands.

%% file: tex/a_experiment_samples.tex
\section{Additional Experimental Details for Section Existing TSQA Benchmarks}
\label{a:existing_benchmark}

\subsection{Data Sampling Process}
In Section \ref{s:existing_benchmarks_sec},
we analyze six existing TSQA 
benchmarks to show 
why they are inadequate for evaluating 
the LLMs performance in answering time-sensitive questions.
We have divided these benchmarks into two distinct groups.
The first group consists of benchmarks developed before LLM knowledge cut-off. These benchmarks are derived from Wikipedia, which are highly susceptible to data contamination. 
This group includes TimeQA, TempReason, and MenatQA.
The second group consists of benchmarks developed after knowledge cut-off. These are proposed as 
free from data contamination. 
This group includes FreshQA, RealtimeQA, and TAQA.
We randomly sampled 1500 questions from 
various splits of these benchmarks. 
For each split, we conducted three rounds of random sampling, 
selecting 50 samples in each round, totaling 150 samples. 
This is to ensure a unbiased selection compared to 
single-round sampling strategies.
Table \ref{t:a_existing_benchmarks_samples} presents the exact number of samples for each benchmark. 

\begin{table}[h!]
\small
\centering
\input{tables/existing_benchmark_samples}
\caption{Existing TSQA benchmark used in empirical experiments in section
~\ref{s:existing_benchmarks_sec}. We randomly sample 1650 questions from six TSQA benchmarks. All the benchmarks (except two specific splits) are developed using Wikipedia and grounded in real-world facts.}
\label{t:a_existing_benchmarks_samples}
\end{table}

\subsection{Details on the Human Reviewing Process}
The primary objective of employing human reviewers 
was to assess to what extent 
time-sensitive questions could be answered 
using the information on the web. 
We enlisted three human reviewers for this task. 
Each reviewer was assigned a total of 150 questions. 
These questions were equally distributed across three distinct datasets: TimeQA, TempReason, and MenatQA.

Each reviewer conducted a direct search of the question on the web using standard Google searches. The task was to browse and review up to a maximum of five articles per question. Here is an illustrative example of reviewing process:
Upon receiving a question such as ``Where did Leo Messi play in 2010?'', each reviewer would search it directly on Google. 
The initial search results typically present 
a variety of sources, including articles and interviews. 
Reviewers then scrutinize the five relevant articles 
of their choice that may explicitly mention Messi’s 
affiliation during the specified year. 
If any article confirms his association with ``FC Barcelona'' 
(the gold answer to the question) in 2010, 
the reviewer marks this question as answer `found' on the web
and records the citation. 

To ensure the reliability of the review process, 
each submission by a reviewer was cross-checked 
by a second reviewer. 
The first reviewer provided citations for the source of the information, which aided in the verification process. 
Given the objective nature of the questions and 
the clear process for determining answers, 
a single round of cross-checking was sufficient for ensuring accuracy, thereby mitigating the need for double annotation. 
The review process revealed that in 88\% to 98.8\% of cases, 
reviewers successfully located answers to the time-sensitive questions through web searches. 

\subsection{Example of Context and Question Altering Approaches.}
\begin{figure*}[!h]
    \centering
    \includegraphics[width=0.96\textwidth]{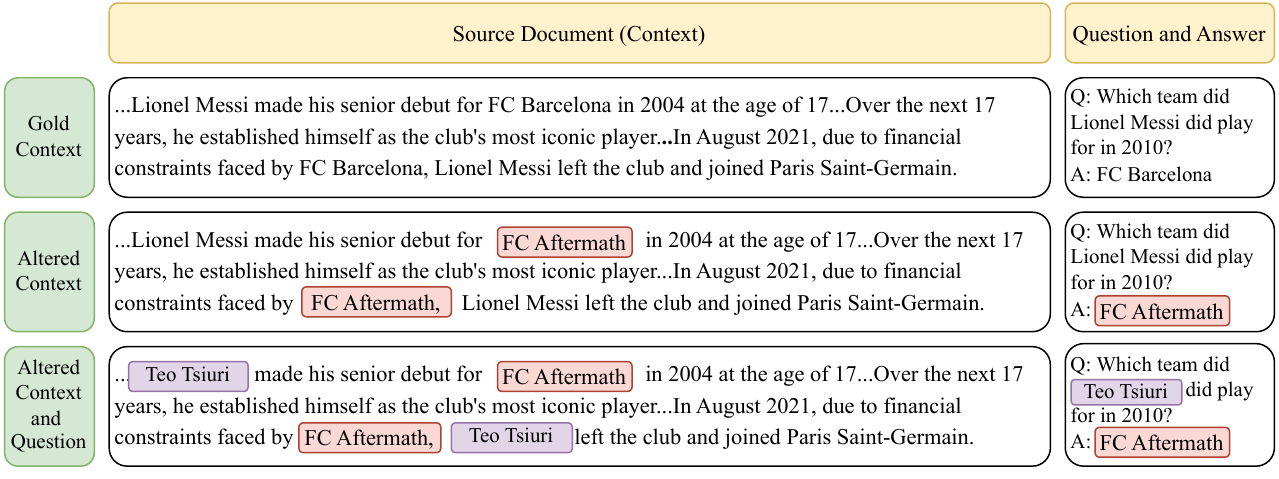}
    \caption{
    Example of different context and question altering approaches. The gold answers are generated from the gold context. In the altered context version, 
    we substitute the original answer with a random entity (e.g., replacing ``FC Barcelona'' with ``FC Aftermath'') while keeping the question unchanged. Additionally, we replace the main entity in the context, ``Lionel Messi'' with ``Teo Tsiuri'' which leads to changes in both the question and the answer.
}
    \label{f:question_alter}
\end{figure*}

We introduce several approaches (in Section \ref{ss:existing_benchmarks_sec_grp_1}) for altering entitites 
to address time-sensitive questions that are not 
grounded in real-world scenarios. 
Figure \ref{f:question_alter} shows an several 
entity-altering approaches (i.e., altered context, altered context and question)
for the TSQA benchmarks.

%% file: tables/existing_benchmark_samples.tex
\begin{tabular}{lcccc}
\toprule
\textbf{Dataset} & \textbf{\#Samples} & \textbf{Source Documents} & \textbf{Time Range} & \textbf{Real-world Facts} \\
\midrule
\textbf{TimeQA~\cite{Chen2021ADF}} & & & & \\
~~Easy               & 150          & Wikipedia                   & 1367 - 2018             & \greentick      \\ 
~~Hard               & 150          & Wikipedia                   & 1367 - 2018             & \greentick      \\
\midrule
\textbf{TempReason~\cite{tan-etal-2023-towards}}    & & & & \\
~~Event-Time         & 150           & Wikipedia                  & 998 - 2023               & \greentick       \\ 
~~Event-Event        & 150           & Wikipedia                  & 998 - 2023               & \greentick       \\
\midrule
\textbf{MenatQA~\cite{wei-etal-2023-menatqa}}       & & & & \\
~~Scope              & 150           & Wikipedia                  & 1367 - 2018              & \greentick      \\ 
~~Order              & 150           & Wikipedia                  & 1367 - 2018              & \greentick        \\ 
~~Counterfactual     & 150           & Wikipedia                  & 1367 - 2018              & \redcross       \\ 
\midrule
\textbf{RealTimeQA~\cite{kasai2023realtime}}  & 150           & News Articles              & 2022 - 2024              & \greentick      \\
\midrule
\textbf{FreshQA~\cite{vu-etal-2024-freshllms}}     & 150           & Wikipedia              & 2022 - Present                       & \greentick      \\
\midrule
\textbf{TAQA~\cite{zhao-etal-2024-set}}        & 150           & Wikipedia                  & 2000 - 2023            & \greentick      \\
\bottomrule
\end{tabular}

%% file: tex/a_additional_results.tex
\section{Experiments with Few-shot Prompting}
\label{a:few_shot_results}

Few-shot prompting often demonstrates superior performance compared to zero-shot prompting across various NLP tasks
\cite{yao-etal-2024-samples, kojima2022large}. 
To assess the model's intrinsic temporal reasoning capabilities, 
we first conducted zero-shot chain-of-thought experiments
in Section \ref{s:experiments}, 
where no additional examples were provided. 
This approach ensures that the model's performance 
is evaluated without any external influence, 
highlighting its inherent ability to reason 
about temporal information.

To further explore the model's capabilities, we extended the experimental setup to few-shot prompting. The key distinction between zero-shot and few-shot prompting is
the inclusion of two randomly selected examples from the UnSeenTimeQA benchmark in the prompt. 
These examples serve as guidance for the model while 
solving the time-sensitive questions. 
We ensured that the selected examples 
did not overlap with the test samples 
to prevent any data leakage or bias in the evaluation. 
We present the few-shot experiment results 
for Easy and Medium question difficulty
in Table \ref{t:few_unseen_benchmarks_1}. 
Additionally, the performance on Hard Serial and Hard Parallel questions is reported in Table \ref{t:few_unseen_benchmarks_2}.

\begin{table*}[h]
\small
\centering
\resizebox{\textwidth}{!}{
\input{tables/few_results_on_easy_and_mid}
}
\caption{Average accuracy and standard deviation across three splits of the easy and medium level questions from the UnSeenTimeQA in few-shot prompting setup. Each split has 300 questions per question types. Overall, result trend is similar to the zero-shot prompting with slight improvements. While large models such as Llama-3.1-70B and GPT-4o have higher performance, the smaller models face challenges in answering simple time-sensitive questions. 
}
\label{t:few_unseen_benchmarks_1}
\end{table*}

\begin{table*}[h]
\small
\centering
\resizebox{\textwidth}{!}{
\input{tables/few_results_on_hard}

}

\caption{Average accuracy and standard deviation across three splits of the hard serial and hard parallel level questions from the UnSeenTimeQA. Each split has 300 questions per question types. Both hard serial and hard parallel level questions are challenging for the models, similar to the findings of zero-shot prompting. Overall, the GPT-4o model performed better than other models yet it missed nearly half of the questions for hard parallel level.}
\label{t:few_unseen_benchmarks_2}
\end{table*}

\newpage

\begin{figure*}[!h]
    \centering
    \includegraphics[width=\textwidth]{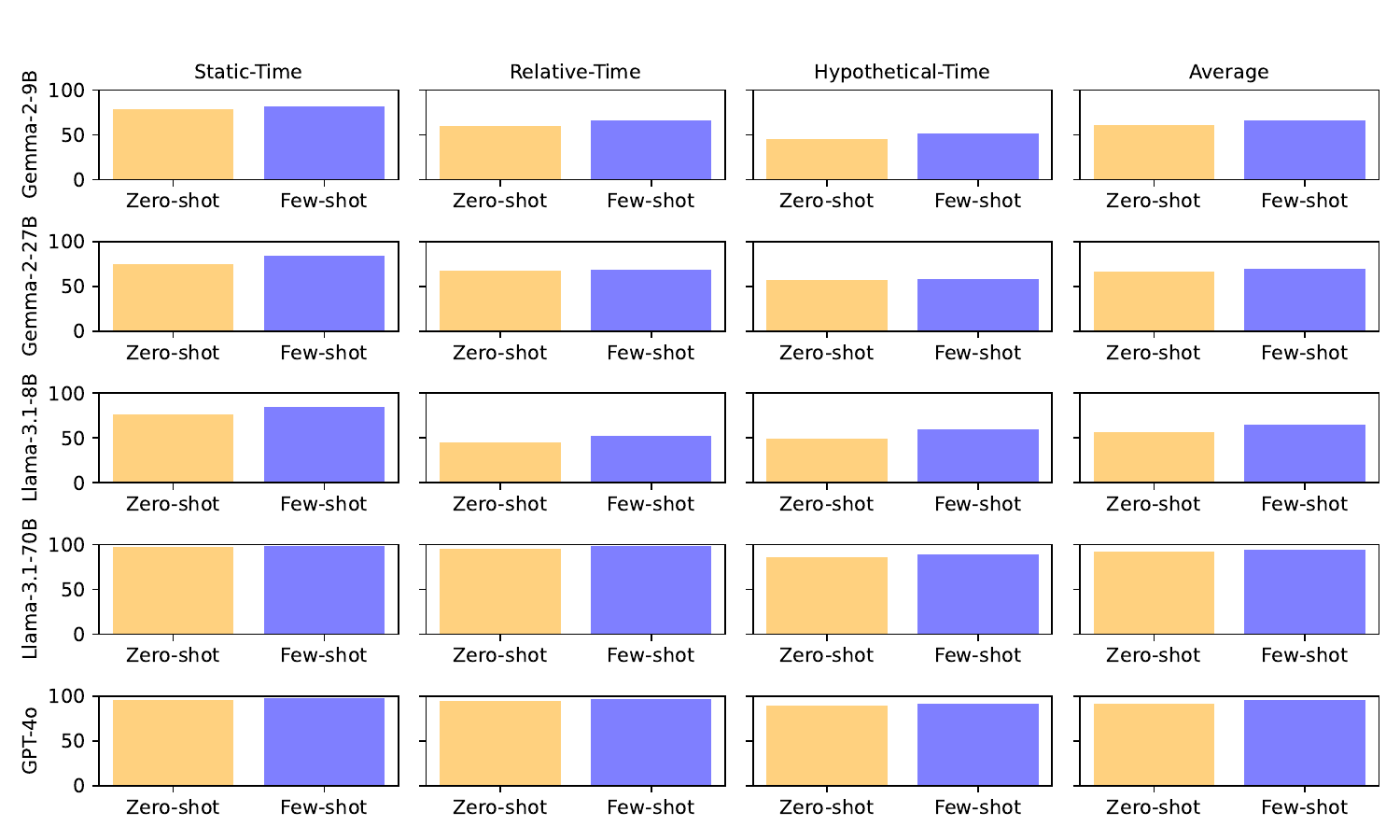}
    \caption{
    Model-wise accuracy comparison between zero- and few-shot prompting for the questions in easy difficulty. Adding example samples in the prompt slightly improved the performance for most of models.  
}
    \label{f:easy_zero_few_shot_compare}
\end{figure*}

\begin{figure*}[!h]
    \centering
    \includegraphics[width=\textwidth]{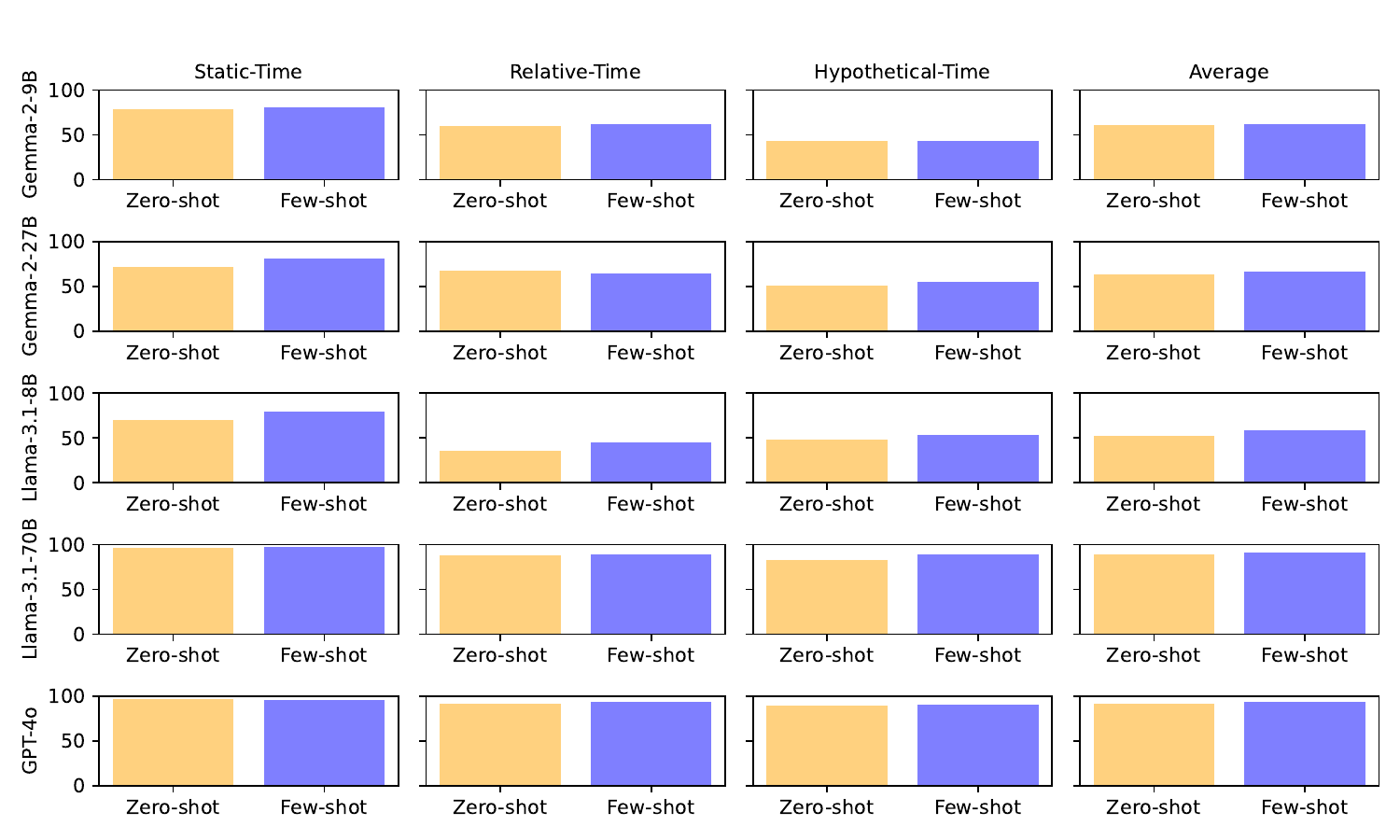}
    \caption{
    Model-wise accuracy comparison between zero- and few-shot prompting for the questions in medium difficulty. Adding example samples in the prompt slightly improved the performance for most of models.  
}
    \label{f:medium_zero_few_shot_compare}
\end{figure*}

\begin{figure*}[!h]
    \centering
    \includegraphics[width=\textwidth]{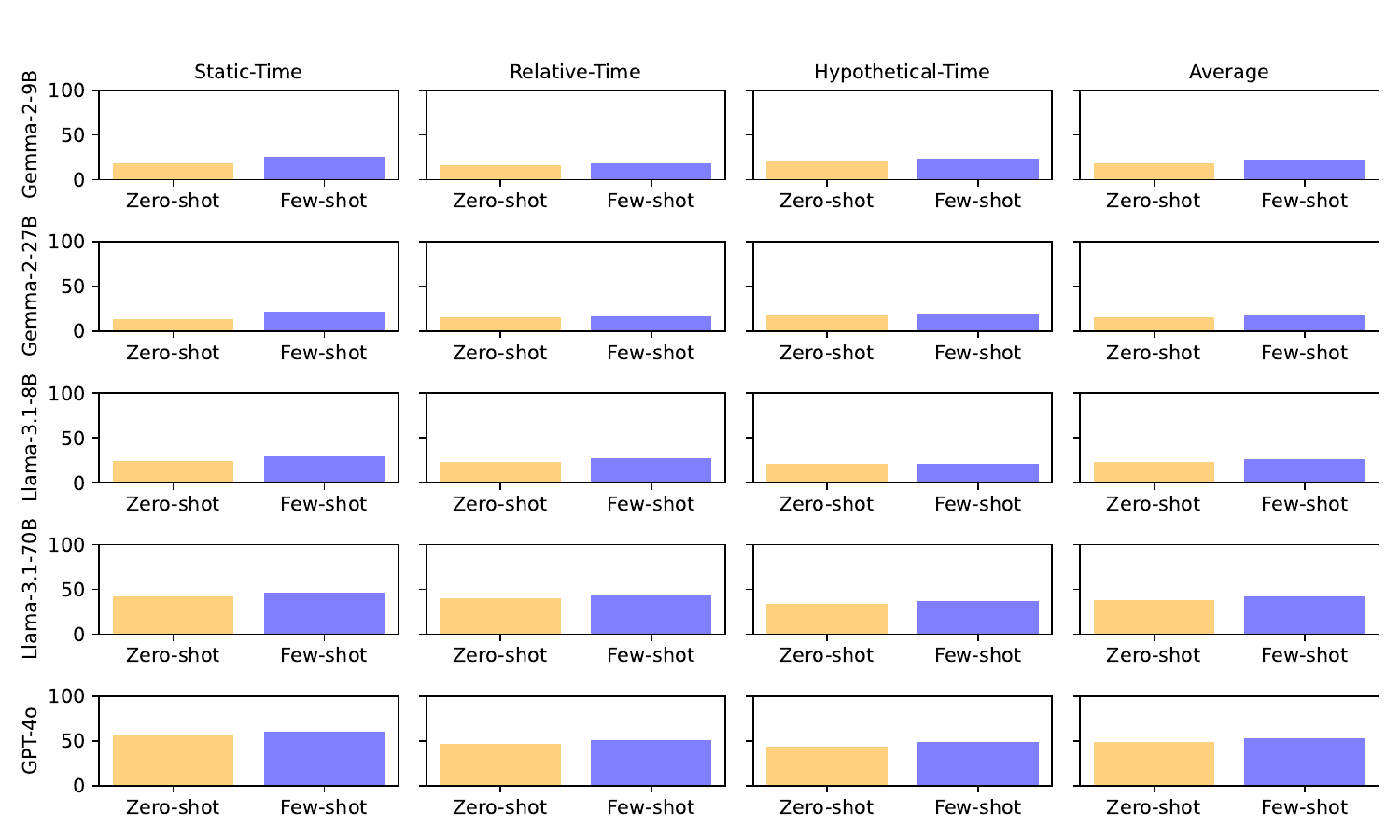}
    \caption{
    Model-wise accuracy comparison between zero- and few-shot prompting for the questions in hard (serial) difficulty. Adding example samples in the prompt slightly improved the performance for most of models.  
}
    \label{f:hard_serial_zero_few_shot_compare}
\end{figure*}

\begin{figure*}[!h]
    \centering
    \includegraphics[width=\textwidth]{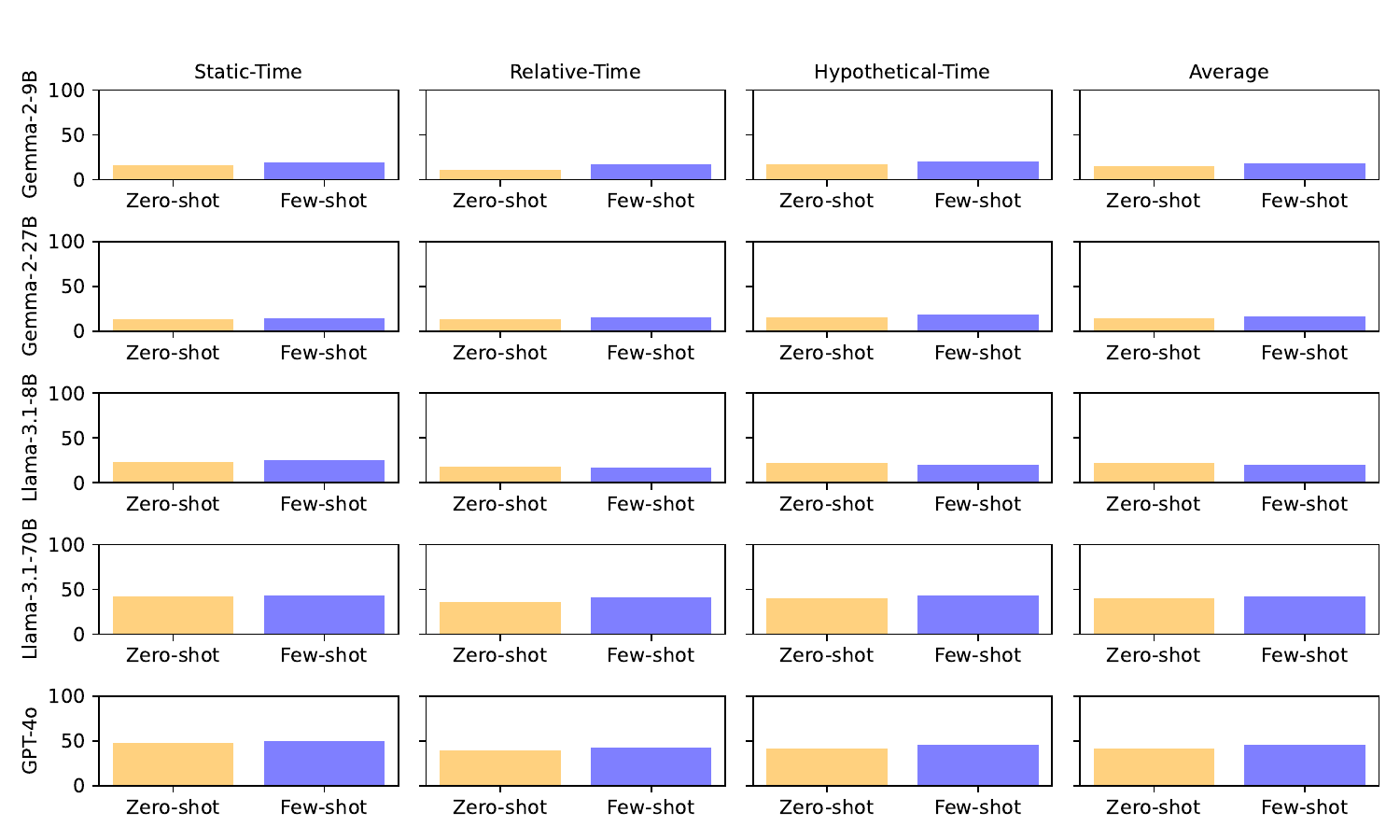}
    \caption{
    Model-wise accuracy comparison between zero- and few-shot prompting for the questions in hard (parallel) difficulty. Adding example samples in the prompt slightly improved the performance for most of models.  
}
    \label{f:hard_parallel_zero_few_shot_compare}
\end{figure*}

Few-shot chain-of-thought prompting enhances model accuracy 
for most of the cases compared to zero-shot prompting. 
Model-wise smaller models like Gemma-2-9B and Llama-3.1-8B exhibit more accuracy improvements than larger models such as Llama-3.1-70B and GPT-4o. 

In figure \ref{f:easy_zero_few_shot_compare},
we compare the accuracy for all models 
for all three question types in easy difficulty. 
Models show improvement in accuracy 
after adding samples in the prompt for all cases. 
On Average, we observed highest increase in accuracy
for the Llama-3.1-8B model. 
While adding samples in the prompt is helpful for larger models,
the accuracy gain is 2.55\% for the Llama-3.1-70B 
and 2.13\% for the GPT-4o.

In figure \ref{f:medium_zero_few_shot_compare},
we compare the accuracy for all models 
for all three question types in medium difficulty. 
Most of the models show accuracy improvement
except 
two scenarios~(Gemma-2-27B model for relative-time questions 
and GPT-4o model for static-time question show accuracy decrease 
after adding samples in the prompt). 
On Average, we observed highest increase in accuracy
for Llama-3.1-8B model. 

In figure \ref{f:hard_serial_zero_few_shot_compare},
we compare the accuracy for all models 
for all three question types in hard (serial) difficulty. 
Models show improvement in accuracy 
after adding samples in the prompt for all cases. 
We observe the highest accuracy improvement for both Gemma models. 
For Gemma-2-9B average improvement is 22\% and 
for Gemma-2-27B average improvement is 31\%. 
The improvement is high because the zero-shot performance is very low.
In Figure \ref{f:hard_parallel_zero_few_shot_compare},
we compare the accuracy for all models 
for all three question types in hard (parallel) difficulty. 
The Gemma models (Gemma-2-9B and Gemma-2-27B)
show higher improvement than the Llama models 
(Llama-3.1-8B and Llama-3.1-70B). 
However, all models demonstrate subpar performance 
on hard parallel difficulty, 
indicating challenges in achieving competitive accuracy.
Even the best model (GPT-4o) still
can not predict correct answers for nearly 50\%
of the time-sensitive question from the 
UnSeenTimeQA benchmark. 

%% file: tables/few_results_on_easy_and_mid.tex
\begin{tabular}{l cccc|cccc} 
\toprule \textbf{Model} & \multicolumn{4}{c}{\textbf{Easy}} & \multicolumn{4}{c}{\textbf{Medium}} \\ 
\cmidrule{2-5} \cmidrule{6-9} & Static-Time & Relative-Time & Hypothetical-Time & Average & Static-Time & Relative-Time & Hypothetical-Time & Average\\ \midrule 

Gemma-2-9B & 82.44\scriptsize{\(\pm\)2.18} & 66.11\scriptsize{\(\pm\)1.68} & 52.33\scriptsize{\(\pm\)2.84} & 66.96 & 80.66\scriptsize{\(\pm\)1.72} & 62.33\scriptsize{\(\pm\)1.57} & 43.22\scriptsize{\(\pm\)1.85} & 62.07 \\ 
Gemma-2-27B & 84.11\scriptsize{\(\pm\)2.17} & 68.88\scriptsize{\(\pm\)2.78} & 58.11\scriptsize{\(\pm\)2.46} & 70.36 & 81.22\scriptsize{\(\pm\)2.23} & 64.22\scriptsize{\(\pm\)2.56} & 54.66\scriptsize{\(\pm\)1.27} & 66.70 \\ 
Llama-3.1-8B & 85.33\scriptsize{\(\pm\)1.59} & 52.77\scriptsize{\(\pm\)0.49} & 59.22\scriptsize{\(\pm\)1.35} & 65.77 & 79.11\scriptsize{\(\pm\)1.76} & 44.77\scriptsize{\(\pm\)4.67} & 53.11\scriptsize{\(\pm\)4.32} & 59.00 \\ 
Llama-3.1-70B & 98.22\scriptsize{\(\pm\)2.65} & 98.68\scriptsize{\(\pm\)1.34} & 89.66\scriptsize{\(\pm\)1.50} & 94.92 & 97.11\scriptsize{\(\pm\)1.42} & 88.77\scriptsize{\(\pm\)1.35} & 88.88\scriptsize{\(\pm\)3.23} & 91.92 \\  
GPT-4o & 97.50\scriptsize{\(\pm\)1.23} & 97.22\scriptsize{\(\pm\)1.76} & 93.22\scriptsize{\(\pm\)1.23} & 95.98 & 95.88\scriptsize{\(\pm\)2.11} & 94.22\scriptsize{\(\pm\)1.28} & 91.33\scriptsize{\(\pm\)2.56} & 93.81 \\ \bottomrule 

\end{tabular}

%% file: tables/few_results_on_hard.tex
\begin{tabular}{l cccc|cccc} 
\toprule \textbf{Model} & \multicolumn{4}{c}{\textbf{Hard (Serial)}} & \multicolumn{4}{c}{\textbf{Hard (Parallel)}} \\ 
\cmidrule{2-5} \cmidrule{6-9} & Static-Time & Relative-Time & Hypothetical-Time & Average & Static-Time & Relative-Time & Hypothetical-Time & Average\\ \midrule 

Gemma-2-9B & 26.33\scriptsize{\(\pm\)2.68} & 18.32\scriptsize{\(\pm\)1.65} & 22.77\scriptsize{\(\pm\)1.41} & 22.47 & 18.50\scriptsize{\(\pm\)1.22} & 16.55\scriptsize{\(\pm\)1.82} & 20.11\scriptsize{\(\pm\)0.54} & 18.38 \\
Gemma-2-27B & 21.55\scriptsize{\(\pm\)1.72} & 17.66\scriptsize{\(\pm\)1.23} & 19.50\scriptsize{\(\pm\)0.88} & 19.57 & 14.44\scriptsize{\(\pm\)1.68} & 16.11\scriptsize{\(\pm\)2.32} & 19.10\scriptsize{\(\pm\)0.89} & 16.55 \\  
Llama-3.1-8B & 29.11\scriptsize{\(\pm\)1.85} & 27.44\scriptsize{\(\pm\)2.78} & 21.33\scriptsize{\(\pm\)1.37} & 25.96 & 25.11\scriptsize{\(\pm\)1.89} & 17.11\scriptsize{\(\pm\)1.11} & 19.55\scriptsize{\(\pm\)1.47} & 20.59 \\ 
Llama-3.1-70B & 46.34\scriptsize{\(\pm\)2.68} & 43.50\scriptsize{\(\pm\)1.32} & 36.50\scriptsize{\(\pm\)1.45} & 42.11 & 43.22\scriptsize{\(\pm\)2.23} & 41.22\scriptsize{\(\pm\)2.55} & 42.64\scriptsize{\(\pm\)1.37} & 42.24 \\  

GPT-4o & 59.56\scriptsize{\(\pm\)0.57} & 51.33\scriptsize{\(\pm\)2.21} & 48.88\scriptsize{\(\pm\)2.52} & 53.25 & 49.50\scriptsize{\(\pm\)2.54} & 42.88\scriptsize{\(\pm\)2.23} & 46.33\scriptsize{\(\pm\)2.22} & 46.23 \\ 
\bottomrule

\end{tabular}

%% file: tex/a_templates.tex
\section{Event Description Templates}
\label{a:templates}

\subsection{Event Templates for Easy Difficulty}
\begin{tcolorbox}[colframe=black, width=\textwidth, boxrule=0.5mm, title=Event Description Templates for Load/Unload Truck and Load/Unload Airplane Event (Easy), breakable]

\begin{compactitem}
    \item at location <location_id>, package <package_id> is \{loaded/unloaded\} into \{truck/airplane\} <vehicle_id> starting at <event_start_time> and finishing at <event_end_time>.
    \item package <package_id> is \{loaded/unloaded\} into \{truck/airplane\} <vehicle_id> from <event_start_time> to <event_end_time> at location <location_id>.
    \item \{loading/unloading\} package <package_id> into \{truck/airplane\} <vehicle_id> at location <location_id> starts at <event_start_time> and ends at <event_end_time>.
    \item from <event_start_time> to <event_end_time> package <package_id> \{loaded/unloaded\} into \{truck/airplane\} <vehicle_id> at location <location_id>.
\end{compactitem}

\end{tcolorbox}

\begin{tcolorbox}[colframe=black, width=\textwidth, boxrule=0.5mm, title=Event Description Templates for Drive Truck Event (Easy), breakable]

\begin{compactitem}
\item from location <start_location_id>, truck <truck_id> moves to location <end_location_id> starting at <event_start_time> and finishing at <event_end_time>.
\item truck <truck_id> operates from location <start_location_id> to location <end_location_id> from <event_start_time> to <event_end_time>.
\item driving truck <truck_id> from location <start_location_id> to location <end_location_id> starts at <event_start_time> and ends at <event_end_time>.
\item from <event_start_time> to <event_end_time> truck <truck_id> transports from location <start_location_id> to location <end_location_id>.
\end{compactitem}

\end{tcolorbox}

\begin{tcolorbox}[colframe=black, width=\textwidth, boxrule=0.5mm, title=Event Description Templates for Fly Airplane Event (Easy), breakable]

\begin{compactitem}
\item from location <start_location_id>, airplane <airplane_id> transits to location <end_location_id> starting at <event_start_time> and finishing at <event_end_time>.
\item airplane <airplane_id> flies from location <start_location_id> to location <end_location_id> from <event_start_time> to <event_end_time>.
\item flying airplane <airplane_id> from location <start_location_id> to location <end_location_id> starts at <event_start_time> and ends at <event_end_time>.
\item from <event_start_time> to <event_end_time> airplane <airplane_id> transits from location <start_location_id> to location <end_location_id>.
\end{compactitem}

\end{tcolorbox}

\subsection{Event Templates for Medium Difficulty}

\begin{tcolorbox}[colframe=black, width=\textwidth, boxrule=0.5mm, title=Event Description Templates for Load/Unload Truck and Load/Unload Airplane Event (Medium), breakable]

\begin{compactitem}
\item at location <location_id>, package <package_id> is \{loaded/unloaded\} into \{truck/airplane\} <vehicle_id> starting at <event_start_time> and continues for \{event_duration\} minutes.
\item package <package_id> is \{loaded/unloaded\} into \{truck/airplane\} <vehicle_id> from <event_start_time> at location <location_id> and takes \{event_duration\} minutes to finish.
\item \{loading/unloading\} package <package_id> into \{truck/airplane\} <vehicle_id> at location <location_id> starts at <event_start_time> and ends after \{event_duration\} minutes.
\item from <event_start_time> package <package_id> is \{loaded/unloaded\} into \{truck/airplane\} <vehicle_id> at location <location_id> for \{event_duration\} minutes.
\end{compactitem}

\end{tcolorbox}

\begin{tcolorbox}[colframe=black, width=\textwidth, boxrule=0.5mm, title=Event Description Templates for Drive Truck Event (Medium), breakable]

\begin{compactitem}
\item from location <start_location_id>, truck <truck_id> moves to location <end_location_id> starting at <event_start_time> and continues for \{event_duration\} minutes.
\item truck <truck_id> operates from location <start_location_id> to location <end_location_id> starting at <event_start_time> and takes \{event_duration\} minutes.
\item driving truck <truck_id> from location <start_location_id> to location <end_location_id> starts at <event_start_time> and ends after \{event_duration\} minutes.
\item from <event_start_time>, truck <truck_id> transports from location <start_location_id> to location <end_location_id> for \{event_duration\} minutes.
\end{compactitem}

\end{tcolorbox}

\begin{tcolorbox}[colframe=black, width=\textwidth, boxrule=0.5mm, title=Event Description Templates for Fly Airplane Event (Medium), breakable]

\begin{compactitem}
\item from location <start_location_id>, airplane <airplane_id> flies to location <end_location_id> starting at <event_start_time> and continues for \{event_duration\} minutes.
\item airplane <airplane_id> flies from location <start_location_id> to location <end_location_id> starting at <event_start_time> and takes \{event_duration\} minutes.
\item flying airplane <airplane_id> from location <start_location_id> to location <end_location_id> starts at <event_start_time> and ends after \{event_duration\} minutes.
\item from <event_start_time>, airplane <airplane_id> transits from location <start_location_id> to location <end_location_id> for \{event_duration\} minutes.
\end{compactitem}

\end{tcolorbox}

\subsection{Event Templates for Hard Difficulty}

\begin{tcolorbox}[colframe=black, width=\textwidth, boxrule=0.5mm, title=Event Description Templates for Load/Unload Truck and Load/Unload Airplane Event (Hard), breakable]

\begin{compactitem}
\item at location <location_id>, package <package_id> is \{loaded/unloaded\} into {truck/airplane} <vehicle_id> and it takes \{event_duration\} minutes to finish.
\item package <package_id> is \{loaded/unloaded\} into \{truck/airplane\} <vehicle_id> at location <location_id> and it requires \{event_duration\} minutes to complete.
\item \{loading/unloading\} package <package_id> into \{truck/airplane\} <vehicle_id> at location <location_id> takes \{event_duration\} minutes to finish.
\item from <event_start_time> to <event_end_time>, package <package_id> is \{loaded/unloaded\} into \{truck/airplane\} <vehicle_id> at location <location_id>.
\end{compactitem}

\end{tcolorbox}

\begin{tcolorbox}[colframe=black, width=\textwidth, boxrule=0.5mm, title=Event Description Templates for Drive Truck Event (Hard), breakable]

\begin{compactitem}
\item from location <start_location_id>, truck <truck_id> moves to location <end_location_id> and it takes \{event_duration\} minutes to finish.
\item truck <truck_id> operates from location <start_location_id> to location <end_location_id> and it requires \{event_duration\} minutes to complete.
\item driving truck <truck_id> from location <start_location_id> to location <end_location_id> takes \{event_duration\} minutes to finish.
\item truck <truck_id> transports from location <start_location_id> to location <end_location_id> for \{event_duration\} minutes.
\end{compactitem}

\end{tcolorbox}

\begin{tcolorbox}[colframe=black, width=\textwidth, boxrule=0.5mm, title=Event Description Templates for Fly Airplane Event (Hard), breakable]

\begin{compactitem}
\item from location <start_location_id>, airplane <airplane_id> transits to location <end_location_id> and it takes \{event_duration\} minutes to finish.
\item airplane <airplane_id> flies from location <start_location_id> to location <end_location_id> and it requires \{event_duration\} minutes to complete.
\item flying airplane <airplane_id> from location <start_location_id> to location <end_location_id> takes \{event_duration\}  minutes to finish.
\item airplane <airplane_id> transits from location <start_location_id> to location <end_location_id> for \{event_duration\} minutes.
\end{compactitem}

\end{tcolorbox}

%% file: tex/a_question_samples.tex
\section{Examples Questions}
\label{a_example_samples}

\subsection{Easy - Static Time}
\begin{tcolorbox}[colframe=black, width=\textwidth, boxrule=0.5mm, title=Example Sample, breakable]

\textbf{[Domain Description]} \\
Loading a package in a truck is possible if the package and the truck are in the same location. During the loading truck event, the package location can be either at the loading location or inside the truck. Loading a package in an airplane is possible if the package and the airplane are in the same location. During the loading airplane event, the package location can be either at the loading location or inside the airplane. Unloading a package from a truck is possible if the package and the truck are in the same location. During the unloading truck event, the package location can be either at the unloading location or inside the truck. Unloading a package from an airplane is possible if the package and the airplane are in the same location. During the unloading airplane event, the package location can be either at the unloading location or inside the airplane. Driving a truck is possible only if the source and destination locations are in the same city. During the driving event, the package location is in the truck. Flying an airplane is possible only if the source and destination locations are in different cities. During the flying event, the package location is in the airplane. Loading and unloading events for any trucks or airplanes, are performed one package at a time. If any event is delayed or expedited, all subsequent events are also delayed or expedited accordingly. \\

\textbf{[Objects Description]} \\
there are 3 cities, c2, c0, and c1. there are 9 locations, l2_1, l1_1, l0_2, l1_2, l2_2, l0_1, l0_0, l2_0, and l1_0. locations l2_1, l2_2, and l2_0 are in city c2. locations l0_2, l0_1, and l0_0 are in city c0. locations l1_1, l1_2, and l1_0 are in city c1. there are 3 airports, The location of the airports are l0_0, l2_0, and l1_0. there are 2 airplanes, a1 and a0. there are 3 trucks, t2, t1, and t0. there are 4 packages, p0, p2, p1, and p3. \\

\textbf{[Initial States Description]} \\
airplane a1 is at the location l1_0. truck t2 is at the location l2_0. truck t1 is at the location l1_1. package p0 is at the location l1_0. package p3 is at the location l1_2. package p2 is at the location l1_1. airplane a0 is at the location l2_0. package p1 is at the location l1_0. truck t0 is at the location l0_0. \\

\textbf{[Events]} \\
Given the initial states, the following events occur: \\
loading package p2 into truck t1 at location l1_1 starts at 01:13 AM and ends at 01:42 AM. from 01:44 AM to 02:57 AM truck t1 transports from location l1_1 to location l1_0. package p1 is loaded into truck t1 from 03:00 AM to 04:47 AM at location l1_0. from location l1_0, truck t1 moves to location l1_2 starting at 04:49 AM and finishing at 06:10 AM. from 06:12 AM to 07:34 AM package p3 loaded into truck t1 at location l1_2. from 07:39 AM to 09:13 AM package p2 unloaded from truck t1 at location l1_2. at location l1_2, package p1 is unloaded from truck t1 starting at 09:19 AM and finishing at 09:54 AM. from 09:56 AM to 11:32 AM truck t1 transports from location l1_2 to location l1_0. unloading package p3 from truck t1 at location l1_0 starts at 11:38 AM and ends at 12:45 PM. package p3 is loaded into airplane a1 from 12:52 PM to 01:12 PM at location l1_0. at location l1_0, package p0 is loaded into airplane a1 starting at 01:17 PM and finishing at 01:47 PM. airplane a1 flys from location l1_0 to location l0_0 from 01:49 PM to 02:57 PM. unloading package p0 from airplane a1 at location l0_0 starts at 03:04 PM and ends at 03:47 PM. from 03:54 PM to 04:21 PM package p0 loaded into truck t0 at location l0_0. truck t0 operates from location l0_0 to location l0_2 from 04:27 PM to 05:10 PM. unloading package p0 from truck t0 at location l0_2 starts at 05:12 PM and ends at 06:48 PM. airplane a1 flys from location l0_0 to location l2_0 from 06:50 PM to 08:08 PM. unloading package p3 from airplane a1 at location l2_0 starts at 08:10 PM and ends at 08:16 PM. loading package p3 into truck t2 at location l2_0 starts at 08:21 PM and ends at 08:44 PM. driving truck t2 from location l2_0 to location l2_1 starts at 08:49 PM and ends at 09:16 PM. at location l2_1, package p3 is unloaded from truck t2 starting at 09:22 PM and finishing at 11:11 PM. \\

\textbf{[Question]} \\
Where is the product p3 at 01:34 PM? \\

\textbf{Answers:} ["l1_0", "a1"] 
\end{tcolorbox}

\subsection{Easy - Relative Time}
\begin{tcolorbox}[colframe=black, width=\textwidth, boxrule=0.5mm, title=Example Sample, breakable]

\textbf{[Domain Description]} \\
Loading a package in a truck is possible if the package and the truck are in the same location. During the loading truck event, the package location can be either at the loading location or inside the truck. Loading a package in an airplane is possible if the package and the airplane are in the same location. During the loading airplane event, the package location can be either at the loading location or inside the airplane. Unloading a package from a truck is possible if the package and the truck are in the same location. During the unloading truck event, the package location can be either at the unloading location or inside the truck. Unloading a package from an airplane is possible if the package and the airplane are in the same location. During the unloading airplane event, the package location can be either at the unloading location or inside the airplane. Driving a truck is possible only if the source and destination locations are in the same city. During the driving event, the package location is in the truck. Flying an airplane is possible only if the source and destination locations are in different cities. During the flying event, the package location is in the airplane. Loading and unloading events for any trucks or airplanes, are performed one package at a time. If any event is delayed or expedited, all subsequent events are also delayed or expedited accordingly. \\

\textbf{[Objects Description]}\\
there are 3 cities, c2, c0, and c1. there are 9 locations, l2_1, l1_1, l0_2, l1_2, l2_2, l0_1, l0_0, l2_0, and l1_0. locations l2_1, l2_2, and l2_0 are in city c2. locations l0_2, l0_1, and l0_0 are in city c0. locations l1_1, l1_2, and l1_0 are in city c1. there are 3 airports, The location of the airports are l0_0, l2_0, and l1_0. there are 2 airplanes, a1 and a0. there are 3 trucks, t2, t1, and t0. there are 4 packages, p0, p2, p1, and p3. \\

\textbf{[Initial States Description]} \\
airplane a1 is at the location l1_0. truck t2 is at the location l2_0. truck t1 is at the location l1_1. package p0 is at the location l1_0. package p3 is at the location l1_2. package p2 is at the location l1_1. airplane a0 is at the location l2_0. package p1 is at the location l1_0. truck t0 is at the location l0_0. \\

\textbf{[Events]}\\
Given the initial states, the following events occur: \\
from 02:53 AM to 03:34 AM package p2 loaded into truck t1 at location l1_1. from location l1_1, truck t1 moves to location l1_0 starting at 03:39 AM and finishing at 05:15 AM. from 05:17 AM to 05:52 AM package p1 loaded into truck t1 at location l1_0. from location l1_0, truck t1 moves to location l1_2 starting at 05:55 AM and finishing at 06:29 AM. loading package p3 into truck t1 at location l1_2 starts at 06:35 AM and ends at 08:40 AM. unloading package p2 from truck t1 at location l1_2 starts at 08:47 AM and ends at 10:10 AM. at location l1_2, package p1 is unloaded from truck t1 starting at 10:13 AM and finishing at 11:45 AM. from location l1_2, truck t1 moves to location l1_0 starting at 11:50 AM and finishing at 01:01 PM. from 01:06 PM to 01:30 PM package p3 unloaded from truck t1 at location l1_0. at location l1_0, package p3 is loaded into airplane a1 starting at 01:33 PM and finishing at 01:38 PM. package p0 is loaded into airplane a1 from 01:40 PM to 02:32 PM at location l1_0. flying airplane a1 from location l1_0 to location l0_0 starts at 02:39 PM and ends at 03:25 PM. at location l0_0, package p0 is unloaded from airplane a1 starting at 03:28 PM and finishing at 05:09 PM. package p0 is loaded into truck t0 from 05:16 PM to 05:19 PM at location l0_0. from location l0_0, truck t0 moves to location l0_2 starting at 05:21 PM and finishing at 06:45 PM. at location l0_2, package p0 is unloaded from truck t0 starting at 06:49 PM and finishing at 06:54 PM. from location l0_0, airplane a1 transits to location l2_0 starting at 06:56 PM and finishing at 08:09 PM. unloading package p3 from airplane a1 at location l2_0 starts at 08:15 PM and ends at 09:35 PM. package p3 is loaded into truck t2 from 09:38 PM to 12:05 AM at location l2_0. from 12:10 AM to 12:22 AM truck t2 transports from location l2_0 to location l2_1. unloading package p3 from truck t2 at location l2_1 starts at 12:29 AM and ends at 12:50 AM. \\

\textbf{[Question]}\\
Where is the package p3 2 hours before 03:50 PM?\\

\textbf{Answers: }["l1_0", "a1"] 

\end{tcolorbox}

\subsection{Easy - Hypothetical Time}
\begin{tcolorbox}[colframe=black, width=\textwidth, boxrule=0.5mm, title=Example Sample, breakable]


\textbf{[Domain Description]} \\
Loading a package in a truck is possible if the package and the truck are in the same location. During the loading truck event, the package location can be either at the loading location or inside the truck. Loading a package in an airplane is possible if the package and the airplane are in the same location. During the loading airplane event, the package location can be either at the loading location or inside the airplane. Unloading a package from a truck is possible if the package and the truck are in the same location. During the unloading truck event, the package location can be either at the unloading location or inside the truck. Unloading a package from an airplane is possible if the package and the airplane are in the same location. During the unloading airplane event, the package location can be either at the unloading location or inside the airplane. Driving a truck is possible only if the source and destination locations are in the same city. During the driving event, the package location is in the truck. Flying an airplane is possible only if the source and destination locations are in different cities. During the flying event, the package location is in the airplane. Loading and unloading events for any trucks or airplanes, are performed one package at a time. If any event is delayed or expedited, all subsequent events are also delayed or expedited accordingly. \\

\textbf{[Objects Description]} \\
there are 3 cities, c0, c2, and c1. there are 6 locations, l0_1, l2_1, l1_1, l1_0, l2_0, and l0_0. locations l0_1 and l0_0 are in city c0. locations l2_1 and l2_0 are in city c2. locations l1_1 and l1_0 are in city c1. there are 3 airports, The location of the airports are l1_0, l2_0, and l0_0. there are 1 airplanes, a0. there are 3 trucks, t1, t0, and t2. there are 6 packages, p1, p3, p4, p0, p2, and p5. \\

\textbf{[Initial States Description]} \\
airplane a0 is at the location l2_0. truck t1 is at the location l1_1. package p2 is at the location l1_1. package p1 is at the location l0_1. truck t0 is at the location l0_0. package p4 is at the location l0_0. package p3 is at the location l2_0. package p5 is at the location l1_0. truck t2 is at the location l2_0. package p0 is at the location l1_0. \\

\textbf{[Events]}\\
Given the initial states, the following events occur: \\
at location l1_1, package p2 is loaded into truck t1 starting at 05:25 AM and finishing at 06:43 AM. from location l1_1, truck t1 moves to location l1_0 starting at 06:49 AM and finishing at 06:54 AM. from 07:00 AM to 07:38 AM package p2 unloaded from truck t1 at location l1_0. from 07:44 AM to 08:59 AM airplane a0 transits from location l2_0 to location l1_0. at location l1_0, package p5 is loaded into airplane a0 starting at 09:06 AM and finishing at 09:46 AM. from 09:51 AM to 10:50 AM package p2 loaded into airplane a0 at location l1_0. at location l1_0, package p0 is loaded into airplane a0 starting at 10:56 AM and finishing at 11:19 AM. airplane a0 flys from location l1_0 to location l0_0 from 11:24 AM to 12:09 PM. at location l0_0, package p5 is unloaded from airplane a0 starting at 12:13 PM and finishing at 12:54 PM. package p5 is loaded into truck t0 from 12:57 PM to 01:13 PM at location l0_0. from 01:15 PM to 02:15 PM truck t0 transports from location l0_0 to location l0_1. from 02:22 PM to 02:33 PM package p5 unloaded from truck t0 at location l0_1. package p1 is loaded into truck t0 from 02:39 PM to 03:03 PM at location l0_1. driving truck t0 from location l0_1 to  xocation l0_0 starts at 03:05 PM and ends at 03:14 PM. at location l0_0, package p1 is unloaded from truck t0 starting at 03:20 PM and finishing at 04:36 PM. package p4 is loaded into airplane a0 from 04:42 PM to 05:38 PM at location l0_0. from 05:40 PM to 06:55 PM package p1 loaded into airplane a0 at location l0_0. from location l0_0, airplane a0 transits to location l2_0 starting at 06:58 PM and finishing at 07:19 PM. from 07:21 PM to 08:05 PM package p3 loaded into airplane a0 at location l2_0. from 08:08 PM to 09:22 PM package p2 unloaded from airplane a0 at location l2_0. unloading package p0 from airplane a0 at location l2_0 starts at 09:24 PM and ends at 09:30 PM. airplane a0 flys from location l2_0 to location l1_0 from 09:36 PM to 10:14 PM. unloading package p4 from airplane a0 at location l1_0 starts at 10:20 PM and ends at 11:42 PM. from 11:49 PM to 12:09 AM package p4 loaded into truck t1 at location l1_0. from 12:11 AM to 12:32 AM package p3 unloaded from airplane a0 at location l1_0. at location l1_0, package p3 is loaded into truck t1 starting at 12:37 AM and finishing at 01:37 AM. unloading package p1 from airplane a0 at location l1_0 starts at 01:42 AM and ends at 03:03 AM. at location l1_0, package p1 is loaded into truck t1 starting at 03:06 AM and finishing at 03:25 AM. from 03:28 AM to 03:31 AM truck t1 transports from location l1_0 to location l1_1. at location l1_1, package p4 is unloaded from truck t1 starting at 03:33 AM and finishing at 03:46 AM. at location l1_1, package p3 is unloaded from truck t1 starting at 03:52 AM and finishing at 04:02 AM. package p1 is unloaded from truck t1 from 04:09 AM to 04:33 AM at location l1_1. \\

\textbf{[Question]}\\
If loading package p5 into airplane a0 at location l1_0 is delayed by 40 minutes, Where is the package p2 at 11:42 AM?\\

\textbf{Answers:} ["l1_0", "a0"]

\end{tcolorbox}

\subsection{Medium - Static Time}
\begin{tcolorbox}[colframe=black, width=\textwidth, boxrule=0.5mm, title=Example Sample, breakable]

\textbf{[Domain Description]}\\
Loading a package in a truck is possible if the package and the truck are in the same location. During the loading truck event, the package location can be either at the loading location or inside the truck. Loading a package in an airplane is possible if the package and the airplane are in the same location. During the loading airplane event, the package location can be either at the loading location or inside the airplane. Unloading a package from a truck is possible if the package and the truck are in the same location. During the unloading truck event, the package location can be either at the unloading location or inside the truck. Unloading a package from an airplane is possible if the package and the airplane are in the same location. During the unloading airplane event, the package location can be either at the unloading location or inside the airplane. Driving a truck is possible only if the source and destination locations are in the same city. During the driving event, the package location is in the truck. Flying an airplane is possible only if the source and destination locations are in different cities. During the flying event, the package location is in the airplane. Loading and unloading events for any trucks or airplanes, are performed one package at a time. If any event is delayed or expedited, all subsequent events are also delayed or expedited accordingly. \\

\textbf{[Objects Description]}\\
there are 3 cities, c0, c2, and c1. there are 6 locations, l0_1, l2_1, l1_1, l1_0, l2_0, and l0_0. locations l0_1 and l0_0 are in city c0. locations l2_1 and l2_0 are in city c2. locations l1_1 and l1_0 are in city c1. there are 3 airports, The location of the airports are l1_0, l2_0, and l0_0. there are 1 airplanes, a0. there are 3 trucks, t1, t0, and t2. there are 6 packages, p1, p3, p4, p0, p2, and p5. \\

\textbf{[Initial States Description]}\\
airplane a0 is at the location l2_0. truck t1 is at the location l1_1. package p2 is at the location l1_1. package p1 is at the location l0_1. truck t0 is at the location l0_0. package p4 is at the location l0_0. package p3 is at the location l2_0. package p5 is at the location l1_0. truck t2 is at the location l2_0. package p0 is at the location l1_0. \\

\textbf{[Question]}\\
Given the initial states, the following events occur: \\
from 07:05 PM package p2 loaded into truck t1 at location l1_1 for 56 minutes. from 08:03 PM truck t1 transports from location l1_1 to location l1_0 for 3 minutes. at location l1_0, package p2 is unloaded from truck t1 starting at 08:08 PM and continues for 62 minutes. flying airplane a0 from location l2_0 to location l1_0 starts at 09:12 PM and ends after 38 minutes. at location l1_0, package p5 is loaded into airplane a0 starting at 09:54 PM and continues for 8 minutes. package p2 is loaded into airplane a0 from 10:05 PM at location l1_0 and takes 67 minutes to finish. package p0 is loaded into airplane a0 from 11:17 PM at location l1_0 and takes 48 minutes to finish. starting at 12:10 AM, airplane a0 flys from location l1_0 to location l0_0 for 3 minutes. unloading package p5 from airplane a0 at location l0_0 starts at 12:20 AM and ends after 49 minutes. at location l0_0, package p5 is loaded into truck t0 starting at 01:13 AM and continues for 17 minutes. starting at 01:37 AM, truck t0 operates from location l0_0 to location l0_1 for 50 minutes. package p5 is unloaded from truck t0 from 02:30 AM at location l0_1 and takes 82 minutes to finish. package p1 is loaded into truck t0 from 03:56 AM at location l0_1 and takes 8 minutes to finish. from 04:07 AM truck t0 transports from location l0_1 to location l0_0 for 22 minutes. from 04:31 AM package p1 unloaded from truck t0 at location l0_0 for 81 minutes. package p4 is loaded into airplane a0 from 05:59 AM at location l0_0 and takes 13 minutes to finish. at location l0_0, package p1 is loaded into airplane a0 starting at 06:19 AM and continues for 37 minutes. from location l0_0, airplane a0 transits to location l2_0 starting at 07:02 AM and continues for 36 minutes. at location l2_0, package p3 is loaded into airplane a0 starting at 07:43 AM and continues for 68 minutes. from 08:57 AM package p2 unloaded from airplane a0 at location l2_0 for 37 minutes. at location l2_0, package p0 is unloaded from airplane a0 starting at 09:39 AM and continues for 21 minutes. starting at 10:04 AM, airplane a0 flys from location l2_0 to location l1_0 for 31 minutes. unloading package p4 from airplane a0 at location l1_0 starts at 10:38 AM and ends after 14 minutes. at location l1_0, package p4 is loaded into truck t1 starting at 10:55 AM and continues for 27 minutes. at location l1_0, package p3 is unloaded from airplane a0 starting at 11:27 AM and continues for 78 minutes. at location l1_0, package p3 is loaded into truck t1 starting at 12:51 PM and continues for 66 minutes. package p1 is unloaded from airplane a0 from 02:00 PM at location l1_0 and takes 23 minutes to finish. package p1 is loaded into truck t1 from 02:28 PM at location l1_0 and takes 43 minutes to finish. from 03:18 PM truck t1 transports from location l1_0 to location l1_1 for 27 minutes. package p4 is unloaded from truck t1 from 03:51 PM at location l1_1 and takes 10 minutes to finish. unloading package p3 from truck t1 at location l1_1 starts at 04:06 PM and ends after 79 minutes. unloading package p1 from truck t1 at location l1_1 starts at 05:28 PM and ends after 43 minutes.\\

\textbf{[Question]}\\
Where is the product p5 at 02:21 AM? \\

\textbf{Answers:} ["t0"]

\end{tcolorbox}

\subsection{Medium - Relative Time}
\begin{tcolorbox}[colframe=black, width=\textwidth, boxrule=0.5mm, title=Example Sample, breakable]

\textbf{[Domain Description]}\\
Loading a package in a truck is possible if the package and the truck are in the same location. During the loading truck event, the package location can be either at the loading location or inside the truck. Loading a package in an airplane is possible if the package and the airplane are in the same location. During the loading airplane event, the package location can be either at the loading location or inside the airplane. Unloading a package from a truck is possible if the package and the truck are in the same location. During the unloading truck event, the package location can be either at the unloading location or inside the truck. Unloading a package from an airplane is possible if the package and the airplane are in the same location. During the unloading airplane event, the package location can be either at the unloading location or inside the airplane. Driving a truck is possible only if the source and destination locations are in the same city. During the driving event, the package location is in the truck. Flying an airplane is possible only if the source and destination locations are in different cities. During the flying event, the package location is in the airplane. Loading and unloading events for any trucks or airplanes, are performed one package at a time. If any event is delayed or expedited, all subsequent events are also delayed or expedited accordingly. \\

\textbf{[Objects Description]}\\
there are 2 cities, c0 and c1. there are 6 locations, l0_1, l0_2, l1_1, l1_2, l1_0, and l0_0. locations l0_1, l0_2, and l0_0 are in city c0. locations l1_1, l1_2, and l1_0 are in city c1. there are 2 airports, The location of the airports are l1_0 and l0_0. there are 2 airplanes, a1 and a0. there are 3 trucks, t1, t2, and t0. there are 4 packages, p0, p3, p1, and p2. \\

\textbf{[Initial States Description]}\\
truck t1 is at the location l1_2. package p3 is at the location l0_2. package p1 is at the location l1_1. airplane a1 is at the location l1_0. package p2 is at the location l0_2. truck t2 is at the location l0_1. truck t0 is at the location l0_2. package p0 is at the location l1_2. airplane a0 is at the location l1_0.\\

\textbf{[Events]}\\
Given the initial states, the following events occur:\\
loading package p0 into truck t1 at location l1_2 starts at 11:00 PM and ends after 12 minutes. from location l1_2, truck t1 moves to location l1_1 starting at 11:14 PM and continues for 53 minutes. from 12:11 AM package p1 loaded into truck t1 at location l1_1 for 4 minutes. from location l1_1, truck t1 moves to location l1_0 starting at 12:20 AM and continues for 32 minutes. from 12:57 AM package p1 unloaded from truck t1 at location l1_0 for 25 minutes. package p0 is unloaded from truck t1 from 01:26 AM at location l1_0 and takes 2 minutes to finish. package p3 is loaded into truck t0 from 01:34 AM at location l0_2 and takes 59 minutes to finish. from 02:38 AM package p2 loaded into truck t0 at location l0_2 for 19 minutes. driving truck t0 from location l0_2 to location l0_0 starts at 03:04 AM and ends after 69 minutes. at location l0_0, package p2 is unloaded from truck t0 starting at 04:15 AM and continues for 95 minutes. loading package p1 into airplane a1 at location l1_0 starts at 05:55 AM and ends after 85 minutes. loading package p0 into airplane a1 at location l1_0 starts at 07:22 AM and ends after 94 minutes. from 09:00 AM airplane a1 transits from location l1_0 to location l0_0 for 62 minutes. package p2 is loaded into airplane a1 from 10:06 AM at location l0_0 and takes 8 minutes to finish. package p1 is unloaded from airplane a1 from 10:19 AM at location l0_0 and takes 70 minutes to finish. package p1 is loaded into truck t0 from 11:31 AM at location l0_0 and takes 77 minutes to finish. at location l0_0, package p0 is unloaded from airplane a1 starting at 12:55 PM and continues for 44 minutes. loading package p0 into truck t0 at location l0_0 starts at 01:41 PM and ends after 84 minutes. from location l0_0, truck t0 moves to location l0_1 starting at 03:08 PM and continues for 43 minutes. package p3 is unloaded from truck t0 from 03:58 PM at location l0_1 and takes 90 minutes to finish. from 05:35 PM truck t0 transports from location l0_1 to location l0_2 for 35 minutes. from 06:17 PM package p1 unloaded from truck t0 at location l0_2 for 30 minutes. at location l0_2, package p0 is unloaded from truck t0 starting at 06:54 PM and continues for 72 minutes. flying airplane a1 from location l0_0 to location l1_0 starts at 08:08 PM and ends after 61 minutes. package p2 is unloaded from airplane a1 from 09:15 PM at location l1_0 and takes 14 minutes to finish. \\

\textbf{[Question]}\\
Where is the package p2 2 hours after 08:11 AM? \\

\textbf{Answers:} ["l0_0", "a1"]

\end{tcolorbox}

\subsection{Medium - Hypothetical Time}
\begin{tcolorbox}[colframe=black, width=\textwidth, boxrule=0.5mm, title=Example Sample, breakable]

\textbf{[Domain Description]}\\
Loading a package in a truck is possible if the package and the truck are in the same location. During the loading truck event, the package location can be either at the loading location or inside the truck. Loading a package in an airplane is possible if the package and the airplane are in the same location. During the loading airplane event, the package location can be either at the loading location or inside the airplane. Unloading a package from a truck is possible if the package and the truck are in the same location. During the unloading truck event, the package location can be either at the unloading location or inside the truck. Unloading a package from an airplane is possible if the package and the airplane are in the same location. During the unloading airplane event, the package location can be either at the unloading location or inside the airplane. Driving a truck is possible only if the source and destination locations are in the same city. During the driving event, the package location is in the truck. Flying an airplane is possible only if the source and destination locations are in different cities. During the flying event, the package location is in the airplane. Loading and unloading events for any trucks or airplanes, are performed one package at a time. If any event is delayed or expedited, all subsequent events are also delayed or expedited accordingly. \\

\textbf{[Objects Description]}\\
there are 3 cities, c0, c2, and c1. there are 6 locations, l0_1, l2_1, l1_1, l1_0, l2_0, and l0_0. locations l0_1 and l0_0 are in city c0. locations l2_1 and l2_0 are in city c2. locations l1_1 and l1_0 are in city c1. there are 3 airports, The location of the airports are l1_0, l2_0, and l0_0. there are 1 airplanes, a0. there are 3 trucks, t1, t0, and t2. there are 6 packages, p1, p3, p4, p0, p2, and p5. \\

\textbf{[Initial States Description]}\\
airplane a0 is at the location l2_0. truck t1 is at the location l1_1. package p2 is at the location l1_1. package p1 is at the location l0_1. truck t0 is at  location l0_0. package p4 is at the location l0_0. package p3 is at the location l2_0. package p5 is at the location l1_0. truck t2 is at the location l2_0. package p0 is at the location l1_0. \\

\textbf{[Events]}\\
Given the initial states, the following events occur:\\
package p2 is loaded into truck t1 from 05:50 PM at location l1_1 and takes 61 minutes to finish. driving truck t1 from location l1_1 to location l1_0 starts at 06:53 PM and ends after 41 minutes. package p2 is unloaded from truck t1 from 07:38 PM at location l1_0 and takes 72 minutes to finish. from location l2_0, airplane a0 transits to location l1_0 starting at 08:57 PM and continues for 51 minutes. loading package p5 into airplane a0 at location l1_0 starts at 09:55 PM and ends after 5 minutes. from 10:02 PM package p2 loaded into airplane a0 at location l1_0 for 14 minutes. at location l1_0, package p0 is loaded into airplane a0 starting at 10:22 PM and continues for 13 minutes. starting at 10:37 PM, airplane a0 flys from location l1_0 to location l0_0 for 48 minutes. at location l0_0, package p5 is unloaded from airplane a0 starting at 11:31 PM and continues for 63 minutes. from 12:40 AM package p5 loaded into truck t0 at location l0_0 for 31 minutes. driving truck t0 from location l0_0 to location l0_1 starts at 01:13 AM and ends after 38 minutes. package p5 is unloaded from truck t0 from 01:57 AM at location l0_1 and takes 10 minutes to finish. at location l0_1, package p1 is loaded into truck t0 starting at 02:12 AM and continues for 39 minutes. starting at 02:55 AM, truck t0 operates from location l0_1 to location l0_0 for 54 minutes. unloading package p1 from truck t0 at location l0_0 starts at 03:52 AM and ends after 49 minutes. loading package p4 into airplane a0 at location l0_0 starts at 04:47 AM and ends after 26 minutes. package p1 is loaded into airplane a0 from 05:20 AM at location l0_0 and takes 61 minutes to finish. flying airplane a0 from location l0_0 to location l2_0 starts at 06:23 AM and ends after 24 minutes. package p3 is loaded into airplane a0 from 06:53 AM at location l2_0 and takes 8 minutes to finish. from 07:07 AM package p2 unloaded from airplane a0 at location l2_0 for 34 minutes. package p0 is unloaded from airplane a0 from 07:48 AM at location l2_0 and takes 43 minutes to finish. from location l2_0, airplane a0 transits to location l1_0 starting at 08:38 AM and continues for 62 minutes. at location l1_0, package p4 is unloaded from airplane a0 starting at 09:47 AM and continues for 42 minutes. from 10:31 AM package p4 loaded into truck t1 at location l1_0 for 10 minutes. from 10:45 AM package p3 unloaded from airplane a0 at location l1_0 for 11 minutes. at location l1_0, package p3 is loaded into truck t1 starting at 11:03 AM and continues for 18 minutes. at location l1_0, package p1 is unloaded from airplane a0 starting at 11:24 AM and continues for 55 minutes. at location l1_0, package p1 is loaded into truck t1 starting at 12:21 PM and continues for 57 minutes. driving truck t1 from location l1_0 to location l1_1 starts at 01:24 PM and ends after 27 minutes. at location l1_1, package p4 is unloaded from truck t1 starting at 01:53 PM and continues for 57 minutes. from 02:53 PM package p3 unloaded from truck t1 at location l1_1 for 66 minutes. at location l1_1, package p1 is unloaded from truck t1 starting at 04:03 PM and continues for 60 minutes. \\

\textbf{[Question]}\\
If driving truck t1 from location l1_1 to location l1_0 is expedited by 4 minutes, Where is the package p2 at 10:23 PM? \\

\textbf{Answers:} ["l1_0", "a0"]

\end{tcolorbox}

\subsection{Hard (Serial) - Static Time}
\begin{tcolorbox}[colframe=black, width=\textwidth, boxrule=0.5mm, title=Example Sample, breakable]

\textbf{[Domain Description]}\\
Loading a package in a truck is possible if the package and the truck are in the same location. During the loading truck event, the package location can be either at the loading location or inside the truck. Loading a package in an airplane is possible if the package and the airplane are in the same location. During the loading airplane event, the package location can be either at the loading location or inside the airplane. Unloading a package from a truck is possible if the package and the truck are in the same location. During the unloading truck event, the package location can be either at the unloading location or inside the truck. Unloading a package from an airplane is possible if the package and the airplane are in the same location. During the unloading airplane event, the package location can be either at the unloading location or inside the airplane. Driving a truck is possible only if the source and destination locations are in the same city. During the driving event, the package location is in the truck. Flying an airplane is possible only if the source and destination locations are in different cities. During the flying event, the package location is in the airplane. Loading and unloading events for any trucks or airplanes, are performed one package at a time. If any event is delayed or expedited, all subsequent events are also delayed or expedited accordingly. \\

\textbf{[Objects Description]}\\
there are 3 cities, c0, c2, and c1. there are 6 locations, l0_1, l2_1, l1_1, l1_0, l2_0, and l0_0. locations l0_1 and l0_0 are in city c0. locations l2_1 and l2_0 are in city c2. locations l1_1 and l1_0 are in city c1. there are 3 airports, The location of the airports are l1_0, l2_0, and l0_0. there are 1 airplanes, a0. there are 3 trucks, t1, t0, and t2. there are 6 packages, p1, p3, p4, p0, p2, and p5. \\

\textbf{[Initial States Description]}\\
airplane a0 is at the location l2_0. truck t1 is at the location l1_1. package p2 is at the location l1_1. package p1 is at the location l0_1. truck t0 is at the location l0_0. package p4 is at the location l0_0. package p3 is at the location l2_0. package p5 is at the location l1_0. truck t2 is at the location l2_0. package p0 is at the location l1_0. \\

\textbf{[Events]}\\
Given the initial states, the following events occur:\\
loading package p2 into truck t1 at location l1_1 takes 14 minutes to finish. truck t1 transports from location l1_1 to location l1_0 for 51 minutes. package p2 is unloaded from truck t1 at location l1_0 and it requires 14 minutes to complete. airplane a0 flys from location l2_0 to location l1_0 for 44 minutes. loading package p5 into airplane a0 at location l1_0 takes 50 minutes to finish. package p2 loaded into airplane a0 at location l1_0 for 22 minutes. at location l1_0, package p0 is loaded into airplane a0 and it takes 15 minutes to finish. flying airplane a0 from location l1_0 to location l0_0 takes 76 minutes to finish. unloading package p5 from airplane a0 at location l0_0 takes 14 minutes to finish. package p5 is loaded into truck t0 at location l0_0 for 56 minutes. from location l0_0, truck t0 moves to location l0_1 and it takes 65 minutes to finish. package p5 unloaded from truck t0 at location l0_1 for 64 minutes. at location l0_1, package p1 is loaded into truck t0 and it takes 7 minutes to finish. from location l0_1, truck t0 moves to location l0_0 and it takes 64 minutes to finish. package p1 unloaded from truck t0 at location l0_0 for 18 minutes. at location l0_0, package p4 is loaded into airplane a0 and it takes 56 minutes to finish. at location l0_0, package p1 is loaded into airplane a0 and it takes 57 minutes to finish. flying airplane a0 from location l0_0 to location l2_0 takes 36 minutes to finish. at location l2_0, package p3 is loaded into airplane a0 and it takes 7 minutes to finish. unloading package p2 from airplane a0 at location l2_0 takes 2 minutes to finish. package p0 is unloaded from airplane a0 at location l2_0 and takes 68 minutes to finish. airplane a0 transits from location l2_0 to location l1_0 for 68 minutes. package p4 is unloaded from airplane a0 at location l1_0 and takes 16 minutes to finish. package p4 is loaded into truck t1 at location l1_0 and it requires 46 minutes to complete. unloading package p3 from airplane a0 at location l1_0 takes 6 minutes to finish. loading package p3 into truck t1 at location l1_0 takes 44 minutes to finish. at location l1_0, package p1 is unloaded from airplane a0 and it takes 7 minutes to finish. loading package p1 into truck t1 at location l1_0 takes 74 minutes to finish. truck t1 transports from location l1_0 to location l1_1 for 11 minutes. at location l1_1, package p4 is unloaded from truck t1 and it takes 28 minutes to finish. package p3 unloaded from truck t1 at location l1_1 for 73 minutes. package p1 is unloaded from truck t1 at location l1_1 and it requires 76 minutes to complete.\\

\textbf{[Question]}\\
If loading package p2 into truck t1 at location l1_1 starts at 06:07 PM, where is the package p2 at 09:22 PM? \\

\textbf{Answers:} ["l1_0", "a0"]

\end{tcolorbox}

\subsection{Hard (Serial) - Relative Time}
\begin{tcolorbox}[colframe=black, width=\textwidth, boxrule=0.5mm, title=Example Sample, breakable]


\textbf{[Domain Description]}\\
Loading a package in a truck is possible if the package and the truck are in the same location. During the loading truck event, the package location can be either at the loading location or inside the truck. Loading a package in an airplane is possible if the package and the airplane are in the same location. During the loading airplane event, the package location can be either at the loading location or inside the airplane. Unloading a package from a truck is possible if the package and the truck are in the same location. During the unloading truck event, the package location can be either at the unloading location or inside the truck. Unloading a package from an airplane is possible if the package and the airplane are in the same location. During the unloading airplane event, the package location can be either at the unloading location or inside the airplane. Driving a truck is possible only if the source and destination locations are in the same city. During the driving event, the package location is in the truck. Flying an airplane is possible only if the source and destination locations are in different cities. During the flying event, the package location is in the airplane. Loading and unloading events for any trucks or airplanes, are performed one package at a time. If any event is delayed or expedited, all subsequent events are also delayed or expedited accordingly. \\

\textbf{[Objects Description]}\\
there are 3 cities, c0, c2, and c1. there are 6 locations, l0_1, l2_1, l1_1, l1_0, l2_0, and l0_0. locations l0_1 and l0_0 are in city c0. locations l2_1 and l2_0 are in city c2. locations l1_1 and l1_0 are in city c1. there are 3 airports, The location of the airports are l1_0, l2_0, and l0_0. there are 1 airplanes, a0. there are 3 trucks, t1, t0, and t2. there are 6 packages, p1, p3, p4, p0, p2, and p5. \\

\textbf{[Initial States Description]}\\
airplane a0 is at the location l2_0. truck t1 is at the location l1_1. package p2 is at the location l1_1. package p1 is at the location l0_1. truck t0 is at the location l0_0. package p4 is at the location l0_0. package p3 is at the location l2_0. package p5 is at the location l1_0. truck t2 is at the location l2_0. package p0 is at the location l1_0. \\

\textbf{[Events]}\\
Given the initial states, the following events occur:\\
package p2 is loaded into truck t1 at location l1_1 and it requires 47 minutes to complete. truck t1 transports from location l1_1 to location l1_0 for 36 minutes. package p2 unloaded from truck t1 at location l1_0 for 48 minutes. airplane a0 transits from location l2_0 to location l1_0 for 39 minutes. package p5 is loaded into airplane a0 at location l1_0 and it requires 2 minutes to complete. at location l1_0, package p2 is loaded into airplane a0 and it takes 5 minutes to finish. package p0 loaded into airplane a0 at location l1_0 for 61 minutes. airplane a0 transits from location l1_0 to location l0_0 for 23 minutes. at location l0_0, package p5 is unloaded from airplane a0 and it takes 47 minutes to finish. at location l0_0, package p5 is loaded into truck t0 and it takes 72 minutes to finish. driving truck t0 from location l0_0 to location l0_1 takes 58 minutes to finish. unloading package p5 from truck t0 at location l0_1 takes 22 minutes to finish. package p1 is loaded into truck t0 at location l0_1 and it requires 19 minutes to complete. truck t0 transports from location l0_1 to location l0_0 for 56 minutes. package p1 is unloaded from truck t0 at location l0_0 and it requires 58 minutes to complete. package p4 loaded into airplane a0 at location l0_0 for 29 minutes. loading package p1 into airplane a0 at location l0_0 takes 22 minutes to finish. flying airplane a0 from location l0_0 to location l2_0 takes 25 minutes to finish. loading package p3 into airplane a0 at location l2_0 takes 35 minutes to finish. unloading package p2 from airplane a0 at location l2_0 takes 38 minutes to finish. package p0 is unloaded from airplane a0 at location l2_0 and takes 31 minutes to finish. airplane a0 transits from location l2_0 to location l1_0 for 53 minutes. package p4 is unloaded from airplane a0 at location l1_0 and takes 33 minutes to finish. package p4 is loaded into truck t1 at location l1_0 and it requires 31 minutes to complete. at location l1_0, package p3 is unloaded from airplane a0 and it takes 6 minutes to finish. at location l1_0, package p3 is loaded into truck t1 and it takes 54 minutes to finish. package p1 unloaded from airplane a0 at location l1_0 for 54 minutes. loading package p1 into truck t1 at location l1_0 takes 25 minutes to finish. from location l1_0, truck t1 moves to location l1_1 and it takes 69 minutes to finish. unloading package p4 from truck t1 at location l1_1 takes 55 minutes to finish. package p3 is unloaded from truck t1 at location l1_1 and it requires 54 minutes to complete. package p1 unloaded from truck t1 at location l1_1 for 42 minutes. \\

\textbf{[Question]}\\
If loading package p5 into airplane a0 at location l1_0 starts at 06:51 AM, where is the package p5 3 hours after 07:24 AM? \\

\textbf{Answers:} ["t0"]

\end{tcolorbox}

\subsection{Hard (Serial) - Hypothetical Time}
\begin{tcolorbox}[colframe=black, width=\textwidth, boxrule=0.5mm, title=Example Sample, breakable]

\textbf{[Domain Description]}\\
Loading a package in a truck is possible if the package and the truck are in the same location. During the loading truck event, the package location can be either at the loading location or inside the truck. Loading a package in an airplane is possible if the package and the airplane are in the same location. During the loading airplane event, the package location can be either at the loading location or inside the airplane. Unloading a package from a truck is possible if the package and the truck are in the same location. During the unloading truck event, the package location can be either at the unloading location or inside the truck. Unloading a package from an airplane is possible if the package and the airplane are in the same location. During the unloading airplane event, the package location can be either at the unloading location or inside the airplane. Driving a truck is possible only if the source and destination locations are in the same city. During the driving event, the package location is in the truck. Flying an airplane is possible only if the source and destination locations are in different cities. During the flying event, the package location is in the airplane. Loading and unloading events for any trucks or airplanes, are performed one package at a time. If any event is delayed or expedited, all subsequent events are also delayed or expedited accordingly. \\

\textbf{[Objects Description]}\\
there are 3 cities, c0, c2, and c1. there are 6 locations, l0_1, l2_1, l1_1, l1_0, l2_0, and l0_0. locations l0_1 and l0_0 are in city c0. locations l2_1 and l2_0 are in city c2. locations l1_1 and l1_0 are in city c1. there are 3 airports, The location of the airports are l1_0, l2_0, and l0_0. there are 1 airplanes, a0. there are 3 trucks, t1, t0, and t2. there are 6 packages, p1, p3, p4, p0, p2, and p5. \\

\textbf{[Initial States Description]}\\
airplane a0 is at the location l2_0. truck t1 is at the location l1_1. package p2 is at the location l1_1. package p1 is at the location l0_1. truck t0 is at the location l0_0. package p4 is at the location l0_0. package p3 is at the location l2_0. package p5 is at the location l1_0. truck t2 is at the location l2_0. package p0 is at the location l1_0. \\

\textbf{[Events]}\\
Given the initial states, the following events occur:\\
package p2 is loaded into truck t1 at location l1_1 for 69 minutes. truck t1 operates from location l1_1 to location l1_0 and it requires 37 minutes to complete. unloading package p2 from truck t1 at location l1_0 takes 32 minutes to finish. airplane a0 transits from location l2_0 to location l1_0 for 9 minutes. loading package p5 into airplane a0 at location l1_0 takes 71 minutes to finish. package p2 loaded into airplane a0 at location l1_0 for 36 minutes. package p0 loaded into airplane a0 at location l1_0 for 48 minutes. flying airplane a0 from location l1_0 to location l0_0 takes 12 minutes to finish. at location l0_0, package p5 is unloaded from airplane a0 and it takes 20 minutes to finish. at location l0_0, package p5 is loaded into truck t0 and it takes 44 minutes to finish. from location l0_0, truck t0 moves to location l0_1 and it takes 23 minutes to finish. package p5 unloaded from truck t0 at location l0_1 for 27 minutes. package p1 is loaded into truck t0 at location l0_1 for 36 minutes. truck t0 transports from location l0_1 to location l0_0 for 32 minutes. unloading package p1 from truck t0 at location l0_0 takes 12 minutes to finish. loading package p4 into airplane a0 at location l0_0 takes 70 minutes to finish. at location l0_0, package p1 is loaded into airplane a0 and it takes 20 minutes to finish. flying airplane a0 from location l0_0 to location l2_0 takes 12 minutes to finish. package p3 loaded into airplane a0 at location l2_0 for 66 minutes. unloading package p2 from airplane a0 at location l2_0 takes 46 minutes to finish. at location l2_0, package p0 is unloaded from airplane a0 and it takes 7 minutes to finish. flying airplane a0 from location l2_0 to location l1_0 takes 59 minutes to finish. package p4 is unloaded from airplane a0 at location l1_0 and takes 25 minutes to finish. loading package p4 into truck t1 at location l1_0 takes 43 minutes to finish. unloading package p3 from airplane a0 at location l1_0 takes 58 minutes to finish. package p3 is loaded into truck t1 at location l1_0 for 66 minutes. package p1 is unloaded from airplane a0 at location l1_0 and takes 74 minutes to finish. package p1 is loaded into truck t1 at location l1_0 for 46 minutes. from location l1_0, truck t1 moves to location l1_1 and it takes 19 minutes to finish. at location l1_1, package p4 is unloaded from truck t1 and it takes 5 minutes to finish. at location l1_1, package p3 is unloaded from truck t1 and it takes 54 minutes to finish. package p1 is unloaded from truck t1 at location l1_1 and it requires 70 minutes to complete. \\

\textbf{[Question]}\\
If loading package p1 into truck t0 at location l0_1 starts at 09:06 AM and driving truck t0 from location l0_1 to location l0_0 is expedited by 15 minutes, where is the package p1 at 12:08 PM? \\

\textbf{Answers:} ["l2_0", "a0"]

\end{tcolorbox}

\subsection{Hard (Parallel) - Static Time}
\begin{tcolorbox}[colframe=black, width=\textwidth, boxrule=0.5mm, title=Example Sample, breakable]


\textbf{[Domain Description]}\\
Loading a package in a truck is possible if the package and the truck are in the same location. During the loading truck event, the package location can be either at the loading location or inside the truck. Loading a package in an airplane is possible if the package and the airplane are in the same location. During the loading airplane event, the package location can be either at the loading location or inside the airplane. Unloading a package from a truck is possible if the package and the truck are in the same location. During the unloading truck event, the package location can be either at the unloading location or inside the truck. Unloading a package from an airplane is possible if the package and the airplane are in the same location. During the unloading airplane event, the package location can be either at the unloading location or inside the airplane. Driving a truck is possible only if the source and destination locations are in the same city. During the driving event, the package location is in the truck. Flying an airplane is possible only if the source and destination locations are in different cities. During the flying event, the package location is in the airplane. Multiple packages can be loaded onto or unloaded from a truck simultaneously, but loading and unloading cannot occur at the same time. Similarly, multiple packages can be loaded or unloaded simultaneously from an airplane, but simultaneous loading and unloading are not permitted. When a truck reaches a new location, unloading of packages must occur before loading new packages. When an airplane arrives at a new location, unloading of packages must occur before loading new packages. If any event is delayed or expedited, all subsequent dependent events are also delayed or expedited accordingly. \\

\textbf{[Objects Description]}\\
there are 3 cities, c0, c2, and c1. there are 6 locations, l0_1, l2_1, l1_1, l1_0, l2_0, and l0_0. locations l0_1 and l0_0 are in city c0. locations l2_1 and l2_0 are in city c2. locations l1_1 and l1_0 are in city c1. there are 3 airports, The location of the airports are l1_0, l2_0, and l0_0. there are 1 airplanes, a0. there are 3 trucks, t1, t0, and t2. there are 6 packages, p1, p3, p4, p0, p2, and p5. \\

\textbf{[Initial States Description]}\\
airplane a0 is at the location l2_0. truck t1 is at the location l1_1. package p2 is at the location l1_1. package p1 is at the location l0_1. truck t0 is at the location l0_0. package p4 is at the location l0_0. package p3 is at the location l2_0. package p5 is at the location l1_0. truck t2 is at the location l2_0. package p0 is at the location l1_0. \\

\textbf{[Events]}\\
Given the initial states, the following events occur:\\
loading package p2 into truck t1 at location l1_1 takes 33 minutes to finish. truck t1 operates from location l1_1 to location l1_0 and it requires 28 minutes to complete. at location l1_0, package p2 is unloaded from truck t1 and it takes 52 minutes to finish. flying airplane a0 from location l2_0 to location l1_0 takes 39 minutes to finish. at location l1_0, package p5 is loaded into airplane a0 and it takes 63 minutes to finish. at location l1_0, package p2 is loaded into airplane a0 and it takes 2 minutes to finish. at location l1_0, package p0 is loaded into airplane a0 and it takes 62 minutes to finish. airplane a0 transits from location l1_0 to location l0_0 for 18 minutes. at location l0_0, package p5 is unloaded from airplane a0 and it takes 49 minutes to finish. at location l0_0, package p5 is loaded into truck t0 and it takes 52 minutes to finish. from location l0_0, truck t0 moves to location l0_1 and it takes 8 minutes to finish. at location l0_1, package p5 is unloaded from truck t0 and it takes 37 minutes to finish. at location l0_1, package p1 is loaded into truck t0 and it takes 59 minutes to finish. from location l0_1, truck t0 moves to location l0_0 and it takes 40 minutes to finish. at location l0_0, package p1 is unloaded from truck t0 and it takes 27 minutes to finish. package p4 is loaded into airplane a0 at location l0_0 and it requires 47 minutes to complete. loading package p1 into airplane a0 at location l0_0 takes 12 minutes to finish. from location l0_0, airplane a0 transits to location l2_0 and it takes 51 minutes to finish. package p2 unloaded from airplane a0 at location l2_0 for 69 minutes. unloading package p0 from airplane a0 at location l2_0 takes 64 minutes to finish. at location l2_0, package p3 is loaded into airplane a0 and it takes 29 minutes to finish. airplane a0 transits from location l2_0 to location l1_0 for 33 minutes. package p4 is unloaded from airplane a0 at location l1_0 and takes 64 minutes to finish. unloading package p3 from airplane a0 at location l1_0 takes 65 minutes to finish. package p1 is unloaded from airplane a0 at location l1_0 and takes 10 minutes to finish. package p4 is loaded into truck t1 at location l1_0 and it requires 16 minutes to complete. at location l1_0, package p3 is loaded into truck t1 and it takes 62 minutes to finish. at location l1_0, package p1 is loaded into truck t1 and it takes 25 minutes to finish. from location l1_0, truck t1 moves to location l1_1 and it takes 60 minutes to finish. at location l1_1, package p4 is unloaded from truck t1 and it takes 14 minutes to finish. at location l1_1, package p3 is unloaded from truck t1 and it takes 8 minutes to finish. package p1 unloaded from truck t1 at location l1_1 for 49 minutes.  \\

\textbf{[Question]}\\
If loading package p2 into truck t1 at location l1_1 starts at 11:18 AM Where is the package p2 at 01:15 PM? \\

\textbf{Answers:} ["l0_0", "a0"]
\end{tcolorbox}

\subsection{Hard (Parallel) - Relative Time}
\begin{tcolorbox}[colframe=black, width=\textwidth, boxrule=0.5mm, title=Example Sample, breakable]

\textbf{[Domain Description]}\\
Loading a package in a truck is possible if the package and the truck are in the same location. During the loading truck event, the package location can be either at the loading location or inside the truck. Loading a package in an airplane is possible if the package and the airplane are in the same location. During the loading airplane event, the package location can be either at the loading location or inside the airplane. Unloading a package from a truck is possible if the package and the truck are in the same location. During the unloading truck event, the package location can be either at the unloading location or inside the truck. Unloading a package from an airplane is possible if the package and the airplane are in the same location. During the unloading airplane event, the package location can be either at the unloading location or inside the airplane. Driving a truck is possible only if the source and destination locations are in the same city. During the driving event, the package location is in the truck. Flying an airplane is possible only if the source and destination locations are in different cities. During the flying event, the package location is in the airplane. Multiple packages can be loaded onto or unloaded from a truck simultaneously, but loading and unloading cannot occur at the same time. Similarly, multiple packages can be loaded or unloaded simultaneously from an airplane, but simultaneous loading and unloading are not permitted. When a truck reaches a new location, unloading of packages must occur before loading new packages. When an airplane arrives at a new location, unloading of packages must occur before loading new packages. If any event is delayed or expedited, all subsequent dependent events are also delayed or expedited accordingly. \\

\textbf{[Objects Description]}\\
there are 3 cities, c0, c2, and c1. there are 6 locations, l0_1, l2_1, l1_1, l1_0, l2_0, and l0_0. locations l0_1 and l0_0 are in city c0. locations l2_1 and l2_0 are in city c2. locations l1_1 and l1_0 are in city c1. there are 3 airports, The location of the airports are l1_0, l2_0, and l0_0. there are 1 airplanes, a0. there are 3 trucks, t1, t0, and t2. there are 6 packages, p1, p3, p4, p0, p2, and p5. \\

\textbf{[Initial States Description]}\\
airplane a0 is at the location l2_0. truck t1 is at the location l1_1. package p2 is at the location l1_1. package p1 is at the location l0_1. truck t0 is at the location l0_0. package p4 is at the location l0_0. package p3 is at the location l2_0. package p5 is at the location l1_0. truck t2 is at the location l2_0. package p0 is at the location l1_0. \\

\textbf{[Events]}\\
Given the initial states, the following events occur:\\
at location l1_1, package p2 is loaded into truck t1 and it takes 37 minutes to finish. truck t1 transports from location l1_1 to location l1_0 for 66 minutes. package p2 unloaded from truck t1 at location l1_0 for 73 minutes. airplane a0 transits from location l2_0 to location l1_0 for 23 minutes. package p5 loaded into airplane a0 at location l1_0 for 65 minutes. loading package p2 into airplane a0 at location l1_0 takes 73 minutes to finish. at location l1_0, package p0 is loaded into airplane a0 and it takes 19 minutes to finish. flying airplane a0 from location l1_0 to location l0_0 takes 82 minutes to finish. at location l0_0, package p5 is unloaded from airplane a0 and it takes 8 minutes to finish. package p5 is loaded into truck t0 at location l0_0 and it requires 39 minutes to complete. truck t0 transports from location l0_0 to location l0_1 for 42 minutes. unloading package p5 from truck t0 at location l0_1 takes 4 minutes to finish. package p1 is loaded into truck t0 at location l0_1 for 81 minutes. driving truck t0 from location l0_1 to location l0_0 takes 23 minutes to finish. package p1 is unloaded from truck t0 at location l0_0 and it requires 54 minutes to complete. loading package p4 into airplane a0 at location l0_0 takes 43 minutes to finish. loading package p1 into airplane a0 at location l0_0 takes 10 minutes to finish. flying airplane a0 from location l0_0 to location l2_0 takes 52 minutes to finish. package p2 is unloaded from airplane a0 at location l2_0 and takes 4 minutes to finish. at location l2_0, package p0 is unloaded from airplane a0 and it takes 43 minutes to finish. at location l2_0, package p3 is loaded into airplane a0 and it takes 4 minutes to finish. flying airplane a0 from location l2_0 to location l1_0 takes 18 minutes to finish. package p4 unloaded from airplane a0 at location l1_0 for 17 minutes. package p3 unloaded from airplane a0 at location l1_0 for 70 minutes. package p1 unloaded from airplane a0 at location l1_0 for 34 minutes. at location l1_0, package p4 is loaded into truck t1 and it takes 60 minutes to finish. package p3 is loaded into truck t1 at location l1_0 and it requires 72 minutes to complete. at location l1_0, package p1 is loaded into truck t1 and it takes 72 minutes to finish. from location l1_0, truck t1 moves to location l1_1 and it takes 44 minutes to finish. package p4 is unloaded from truck t1 at location l1_1 and it requires 2 minutes to complete. package p3 unloaded from truck t1 at location l1_1 for 7 minutes. package p1 unloaded from truck t1 at location l1_1 for 7 minutes. \\

\textbf{[Question]}\\
If loading package p2 into truck t1 at location l1_1 starts at 10:09 PM, where is the package p2 3 hours after 10:18 PM? \\

\textbf{Answers:} ["l0_0", "a0"]

\end{tcolorbox}

\subsection{Hard (Parallel) - Hypothetical Time}
\begin{tcolorbox}[colframe=black, width=\textwidth, boxrule=0.5mm, title=Example Sample, breakable]

\textbf{[Domain Description]}\\
Loading a package in a truck is possible if the package and the truck are in the same location. During the loading truck event, the package location can be either at the loading location or inside the truck. Loading a package in an airplane is possible if the package and the airplane are in the same location. During the loading airplane event, the package location can be either at the loading location or inside the airplane. Unloading a package from a truck is possible if the package and the truck are in the same location. During the unloading truck event, the package location can be either at the unloading location or inside the truck. Unloading a package from an airplane is possible if the package and the airplane are in the same location. During the unloading airplane event, the package location can be either at the unloading location or inside the airplane. Driving a truck is possible only if the source and destination locations are in the same city. During the driving event, the package location is in the truck. Flying an airplane is possible only if the source and destination locations are in different cities. During the flying event, the package location is in the airplane. Multiple packages can be loaded onto or unloaded from a truck simultaneously, but loading and unloading cannot occur at the same time. Similarly, multiple packages can be loaded or unloaded simultaneously from an airplane, but simultaneous loading and unloading are not permitted. When a truck reaches a new location, unloading of packages must occur before loading new packages. When an airplane arrives at a new location, unloading of packages must occur before loading new packages. If any event is delayed or expedited, all subsequent dependent events are also delayed or expedited accordingly. \\

\textbf{[Objects Description]}\\
there are 3 cities, c0, c2, and c1. there are 6 locations, l0_1, l2_1, l1_1, l1_0, l2_0, and l0_0. locations l0_1 and l0_0 are in city c0. locations l2_1 and l2_0 are in city c2. locations l1_1 and l1_0 are in city c1. there are 3 airports, The location of the airports are l1_0, l2_0, and l0_0. there are 1 airplanes, a0. there are 3 trucks, t1, t0, and t2. there are 6 packages, p1, p3, p4, p0, p2, and p5. \\

\textbf{[Initial States Description]}\\
airplane a0 is at the location l2_0. truck t1 is at the location l1_1. package p2 is at the location l1_1. package p1 is at the location l0_1. truck t0 is at the location l0_0. package p4 is at the location l0_0. package p3 is at the location l2_0. package p5 is at the location l1_0. truck t2 is at the location l2_0. package p0 is at the location l1_0. \\

\textbf{[Events]}\\
Given the initial states, the following events occur:\\
loading package p2 into truck t1 at location l1_1 takes 60 minutes to finish. truck t1 operates from location l1_1 to location l1_0 and it requires 5 minutes to complete. at location l1_0, package p2 is unloaded from truck t1 and it takes 51 minutes to finish. from location l2_0, airplane a0 transits to location l1_0 and it takes 59 minutes to finish. package p5 loaded into airplane a0 at location l1_0 for 27 minutes. package p2 loaded into airplane a0 at location l1_0 for 8 minutes. at location l1_0, package p0 is loaded into airplane a0 and it takes 6 minutes to finish. airplane a0 transits from location l1_0 to location l0_0 for 16 minutes. unloading package p5 from airplane a0 at location l0_0 takes 67 minutes to finish. at location l0_0, package p5 is loaded into truck t0 and it takes 54 minutes to finish. driving truck t0 from location l0_0 to location l0_1 takes 2 minutes to finish. package p5 unloaded from truck t0 at location l0_1 for 56 minutes. at location l0_1, package p1 is loaded into truck t0 and it takes 22 minutes to finish. from location l0_1, truck t0 moves to location l0_0 and it takes 35 minutes to finish. unloading package p1 from truck t0 at location l0_0 takes 59 minutes to finish. package p4 loaded into airplane a0 at location l0_0 for 53 minutes.package p1 is loaded into airplane a0 at location l0_0 and it requires 24 minutes to complete. from location l0_0, airplane a0 transits to location l2_0 and it takes 21 minutes to finish. package p2 unloaded from airplane a0 at location l2_0 for 68 minutes. at location l2_0, package p0 is unloaded from airplane a0 and it takes 9 minutes to finish. loading package p3 into airplane a0 at location l2_0 takes 42 minutes to finish. airplane a0 flys from location l2_0 to location l1_0 for 59 minutes. unloading package p4 from airplane a0 at location l1_0 takes 50 minutes to finish. package p3 unloaded from airplane a0 at location l1_0 for 32 minutes. at location l1_0, package p1 is unloaded from airplane a0 and it takes 62 minutes to finish. package p4 is loaded into truck t1 at location l1_0 and it requires 11 minutes to complete. at location l1_0, package p3 is loaded into truck t1 and it takes 43 minutes to finish. at location l1_0, package p1 is loaded into truck t1 and it takes 13 minutes to finish. truck t1 transports from location l1_0 to location l1_1 for 61 minutes. package p4 is unloaded from truck t1 at location l1_1 and it requires 78 minutes to complete. package p3 is unloaded from truck t1 at location l1_1 and it requires 18 minutes to complete. package p1 unloaded from truck t1 at location l1_1 for 80 minutes. \\

\textbf{[Question]}\\
If loading package p2 into truck t1 at location l1_1 starts at 06:43 PM and flying airplane a0 from location l2_0 to location l1_0 is delayed by 78 minutes, where is the package p2 at 09:02 PM? \\

\textbf{Answers:} ["l1_0", "a0"]

\end{tcolorbox}

%% file: tex/a_prompts.tex
\section{Prompt for Evaluating UnSeenTimeQA}
\label{a:experiment_prompt}

\begin{tcolorbox}[colback=white, colframe=black, width=\textwidth, boxrule=0.5mm]

The structure of prompts used in the UnSeenTimeQA benchmark is as follows:

\begin{dmath*}
\text{[domain\_description]} + \text{[object\_description]} + \text{[initial\_states\_description]} + \text{[events]} + \text{[question]} + \text{[reasoning\_prompt]}
\end{dmath*}

\begin{itemize}
  \item \textbf{[domain\_description]}: Provides a comprehensive description of the environment, outlining how different events can occur with various objects. 
  
  \item \textbf{[object\_description]}: Lists and describes all relevant objects within the scenario. This includes details such as locations, vehicles, and packages.
  
  \item \textbf{[initial\_states\_description]}: Describes the initial states (mostly locations) of all objects. 
  
  \item \textbf{[events]}: Provide a chronological account of the events from the initial state to the goal state. This should include the movements, actions, and changes of objects over time within the logistics environment, helping to track key developments and transitions.

  \item \textbf{[question]}: A specific query about the state of a package at a given point in time. This requires the model to synthesize the information from the previous sections to provide an accurate answer.
  
  \item \textbf{[reasoning\_prompt]}: Instructs the model to think step-by-step to answer the question, guiding it to generate reasoning steps and a final answer. This helps in structuring the model's response systematically. \\We use this exact prompt: \emph{Let's think step-by-step to answer the question. Please use the below format:\\Reasoning steps: [generate step-by-step reasoning]\\Answer: [final answer]}
\end{itemize}

\end{tcolorbox}

%% file: tex/a_model_details.tex
\section{Model Details}
\label{a:model_details}
\subsection{Model Endpoints}

Table \ref{t:a_model_endpoints} lists the open-weight and close-weight models used in the experiments, along with their corresponding Huggingface repositories and API endpoints.

\begin{table}[h!]
\begin{tabular}{p{5cm}p{10cm}}
\hline
\textbf{Model Endpoint}  & \textbf{Link}                                               \\ \hline
GPT-4o                                & \href{https://platform.openai.com/docs/models/gpt-4o}{https://platform.openai.com/docs/models/gpt-4o} \\
google/gemma-2-9b-it                      & \href{https://huggingface.co/google/gemma-2-9b-it}{https://huggingface.co/google/gemma-2-9b-it}                      \\
google/gemma-2-27b-it                      & \href{https://huggingface.co/google/gemma-2-27b-it}{https://huggingface.co/google/gemma-2-27b-it}                      \\
meta-llama/Meta-Llama-3.1-8B-Instruct  & \href{https://huggingface.co/meta-llama/Meta-Llama-3.1-8B-Instruct}{https://huggingface.co/meta-llama/Meta-Llama-3.1-8B-Instruct}  \\
meta-llama/Meta-Llama-3.1-70B-Instruct & \href{https://huggingface.co/meta-llama/Meta-Llama-3.1-70B-Instruct}{https://huggingface.co/meta-llama/Meta-Llama-3.1-70B-Instruct} \\ \hline
\end{tabular}
\caption{List of open- and close-weight models and their corresponding Huggingface and API endpoints}
\label{t:a_model_endpoints}
\end{table}



\subsection{GPU Hours}
We conducted our experiments using various configurations of A100 GPUs. To run the LLaMA-3.1-70B model, we utilized 24 A100 80GB GPUs, completing all queries across different categories in approximately three days. For the experiments involving the Gemma-2 (9B), LLaMA-3.1 (8B), and Gemma-2 (27B) models, we used 8 A100 80GB GPUs, which took about two days to complete.

%% file: tex/a_dataset_datacard.tex
\section{UnSeenTimeQA Documentation: Datasheet}
In this section, we address benchmark-related questions by following to the guidelines in \newcite{gebru2021datasheets}, to ensure comprehensive documentation
for the benchmark creation, structure, and use.
\subsection{Motivation}
\begin{itemize}
    \item \textbf{For what purpose was the dataset created?}
    \\The dataset was created to benchmark and evaluate large language models (LLMs) on time-sensitive question-answering tasks, with a specific focus on ensuring no data contamination from previously encountered sources. The primary objective is to establish a reliable, contamination-free evaluation benchmark that enables robust assessment of models' temporal reasoning capabilities.
    \item \textbf{Who created the dataset (e.g., which team, research group) and on behalf of which entity (e.g., company, institution, organization)?}
    \\The UnSeenTimeQA dataset was created by the collaborators from the Arizona State University, University of Arizona and New Mexico State University. The project was led by Md Nayem Uddin and advised by Eduardo Blanco and Chitta Baral.
    \item \textbf{Who funded the creation of the dataset?}
    \\The development of UnSeenTimeQA was support by the U.S. Office of Naval Research.
\end{itemize}

\subsection{Composition}
\begin{itemize}
    \item \textbf{What do the instances that comprise the dataset represent (e.g., documents, photos, people, countries)?}
    \\Instances are text only. Each instance represents a time-sensitive question within the logistics domain.
    \item \textbf{How many instances are there in total (of each type, if appropriate)?}
    \\The dataset consists of a total of 10,800 instances. These are divided evenly across four difficulty categories: easy, medium, hard-serial, and hard-parallel, with 2,700 questions in each category. Additionally, the dataset includes three types of questions: static-time, relative-time, and hypothetical-time. Each question type contains 900 instances per difficulty category, resulting in an equal distribution across all combinations of difficulty and question type.
    \item \textbf{Does the dataset contain all possible instances or is it a sample (not necessarily random) of instances from a larger set?}
    \\The dataset consists of synthetically created instances. A large pool of synthetic questions was initially generated, and the final splits were randomly sampled from this pool based on a depth metric. The depth ranges from 6 to 20, with 20 questions sampled for each depth level.
    \item \textbf{What data does each instance consist of?}
    \\Each instance comprises the following components: a domain description, an object description, initial states, a sequence of events, a time-sensitive question, and its corresponding answer.
    \item \textbf{Is there a label or target associated with each instance?}
    \\Yes, each instance includes a target label.
    \item \textbf{Is any information missing from individual instances?}
    \\No, all instances contain the required information, including the target labels
    \item \textbf{Are relationships between individual instances made explicit (e.g., users’ movie ratings, social network links)?}
    \\No, there are no relationships between different instances.
    \item \textbf{Are there recommended data splits (e.g., training, development/validation, testing)?}
    \\No, we propose UnSeenTimeQA solely as an evaluation benchmark and do not include predefined splits for training.
    \item \textbf{Are there any errors, sources of noise, or redundancies in the dataset?}
    \\The dataset was created automatically and multiple rounds of manual validation were conducted to ensure it is free of errors to the best of our knowledge.
    \item \textbf{Is the dataset self-contained, or does it link to or otherwise rely on external resources (e.g., websites, tweets, other datasets)?}
    \\Yes, the dataset is entirely self-contained and does not link to or rely on any external resources.
    \item \textbf{Does the dataset contain data that might be considered confidential (e.g., data that is protected by legal privilege or by doctor–patient confidentiality, data that includes the content of individuals’ non-public communications)?}
    \\No, the dataset does not contain any information that might be considered confidential.
    \item \textbf{Does the dataset contain data that, if viewed directly, might be offensive, insulting, threatening, or might otherwise cause anxiety?}
    \\No, the dataset does not include any content that could be considered offensive, insulting, threatening, or anxiety-inducing.
    \item \textbf{Does the dataset identify any subpopulations (e.g., by age, gender)?}
    \\No, the dataset does not contain any attributes or information that could identify or infer subpopulations.
    \item \textbf{Is it possible to identify individuals (i.e., one or more natural persons), either directly or indirectly (i.e., in combination with other data) from the dataset?}
    \\No, the dataset does not contain any personal information.
    \item \textbf{Does the dataset contain data that might be considered sensitive
    in any way (e.g., data that reveals race or ethnic origins, sexual orientations, religious beliefs, political opinions or union memberships, or locations; financial or health data; biometric or genetic
    data; forms of government identification, such as social security
    numbers; criminal history)?}
    \\No, the dataset does not contain any data that might be considered sensitive in any of these ways.
\end{itemize}

\subsection{Collection}
\begin{itemize}
    \item \textbf{How was the data associated with each instance acquired?}
    \\The data instances were acquired using a Python script specifically designed for the logistics planning domain.
    \item \textbf{What mechanisms or procedures were used to collect the data (e.g., hardware apparatuses or sensors, manual human curation, software programs, software APIs)?}
    \\The data collection process was entirely automated, utilizing custom-developed scripts created by the authors. To ensure data quality and accuracy, multiple rounds of manual validation were conducted to identify and correct potential errors. Further details regarding the data collection mechanisms and validation procedures are provided in the paper.
    \item \textbf{If the dataset is a sample from a larger set, what was the sampling strategy (e.g., deterministic, probabilistic with specific sampling probabilities)?}
    \\Final dataset was curated by evenly distributing instances across a depth metric ranging from 6 to 20, ensuring a balanced representation of different levels of complexity.
    \item \textbf{Who was involved in the data collection process (e.g., students, crowdworkers, contractors) and how were they compensated (e.g., how much were crowdworkers paid)?}
    \\The authors were solely responsible for the data collection process. No external individuals or groups were involved.
    \item \textbf{Over what timeframe was the data collected?}
    \\The data instances were generated and curated using Python scripts. The most recent version of the dataset was finalized and collected in October 2024.
    \item \textbf{Were any ethical review processes conducted (e.g., by an institutional review board)?}
    \\No formal ethical reviews were conducted, as the dataset does not contain any sensitive, personal, or harmful information.
    \item \textbf{Did you collect the data from the individuals in question directly, or obtain it via third parties or other sources (e.g., websites)?}
    \\The data was not collected directly from individuals or obtained through third-party sources.
\end{itemize}

\subsection{Uses}
\begin{itemize}
    \item \textbf{Has the dataset been used for any tasks already?}
    \\The dataset has been used in this paper to evaluate the temporal reasoning capabilities of large language models (LLMs) in time-sensitive question answering. It also serves as a data-contamination-free benchmark to ensure robust and reliable assessment of LLM performance in handling temporal reasoning tasks.
    \item \textbf{Is there a repository that links to any or all papers or systems that use the dataset?}
    \\This is the first paper to use the dataset. However, we plan to create and maintain a repository in the future that links to all papers, systems, and projects utilizing the dataset 
    \item \textbf{What (other) tasks could the dataset be used for?}
    \\The dataset is well-suited for tasks involving temporal reasoning, such as event ordering, temporal relationship extraction, time-sensitive question answering, and timeline reconstruction. Additionally, it can serve as a resource for research in areas like narrative understanding, temporal information retrieval, and other applications requiring nuanced temporal context. 
    \item \textbf{Is there anything about the composition of the dataset or the way it was collected and preprocessed/cleaned/labeled that might impact future uses?}
    \\No.
    \item \textbf{Are there tasks for which the dataset should not be used?}
    \\No.
\end{itemize}

\subsection{Distribution}
\begin{itemize}
    \item \textbf{Will the dataset be distributed to third parties outside of the entity (e.g., company, institution, organization) on behalf of which the dataset was created?}
    \\The dataset will be publicly released on GitHub and Hugging Face.
    \item \textbf{How will the dataset be distributed (e.g., tarball on website, API, GitHub)?}
    \\The dataset will be distributed through a GitHub repository and made accessible via the Hugging Face API, ensuring ease of access and usability for a wide range of users.
    \item \textbf{When will the dataset be distributed?}
    \\The dataset is publicly available.
    \item \textbf{Will the dataset be distributed under a copyright or other intellectual property (IP) license, and/or under applicable terms of use (ToU)?}
    \\The dataset will be released under the Creative Commons Attribution 4.0 International (CC BY 4.0) license. This license allows users to share, adapt, and build upon the dataset for any purpose, including commercial use, as long as appropriate credit is given, any changes are indicated, and the terms of the license are followed. 
    \item \textbf{Have any third parties imposed IP-based or other restrictions on the data associated with the instances?}
    \\No.
    \item \textbf{Do any export controls or other regulatory restrictions apply to the dataset or to individual instances?}
    \\No.
\end{itemize}

\subsection{Maintenance}
\begin{itemize}
    \item \textbf{Who will be supporting/hosting/maintaining the dataset?}
    \\Md Nayem Uddin will be responsible for future maintance of the dataset.
    \item \textbf{How can the owner/curator/manager of the dataset be contacted (e.g., email address)?}
    \\Please raise an issue in the official Github repository.
    \item \textbf{Is there an erratum?}
    \\If any errors are identified in the future, we will update the dataset accordingly and release a revised version, ensuring that all changes are documented and acknowledged.
    \item \textbf{Will the dataset be updated (e.g., to correct labeling errors, add new instances, delete instances)?}
    \\The dataset has been carefully curated to ensure accuracy and reliability. However, if any unforeseen issues arise that require updates, such as refining data labels, adding new instances, or removing specific samples, updates will be made as needed to maintain the dataset's quality.
    \item \textbf{If the dataset relates to people, are there applicable limits on the retention of the data associated with the instances (e.g., were the individuals in question told that their data would be retained for a fixed period of time and then deleted)?}
    \\This dataset does not contain any information that is specific to individuals.
    \item \textbf{Will older versions of the dataset continue to be supported/hosted/maintained?}
    \\All versions of the dataset will remain accessible through the GitHub repository, ensuring that users can access and reference previous versions as needed.
    \item \textbf{If others want to extend/augment/build on/contribute to the dataset, is there a mechanism for them to do so?}
    \\We value any kind of contributions that build upon our dataset. The dataset is publicly available, and anyone is welcome to use it for their research, extend it, augment it, or modify it based on their needs. We encourage collaboration and innovation using our data and look forward to seeing how others enhance and expand upon this work.
\end{itemize}